\documentclass[journal]{IEEEtran}
    
            \usepackage{cite}
    \ifCLASSINFOpdf
                                        \else
                                                          \fi
                                                                                
    \usepackage{graphicx}
        \usepackage{subcaption}
    \usepackage{epstopdf}
    
            \usepackage{amsmath}
    \usepackage{amssymb}
    \usepackage{amsfonts}
    \usepackage{mathtools}
    \usepackage{bbm}
    \usepackage{balance}

    \usepackage{fancyref}

            \usepackage{url}
    \usepackage{color}
    \usepackage{algorithm}
\usepackage[noend]{algpseudocode}

    \usepackage{enumitem}
     \newcommand{\E}{\mathbb{E}}

        \newcommand{\BC}{\boldsymbol{{C}}}
        
        \newcommand{\R}{\boldsymbol{\mathcal{R}}}
        \newcommand{\w}{\boldsymbol{w}}
        \newcommand{\uki}{\boldsymbol{u}_{k,i}}
        \newcommand{\N}{\textit{N}}
        \newcommand{\Nk}{\mathcal{N}_k}

        \newcommand{\bpsi}{\boldsymbol{\psi}}
        \newcommand{\tw}{\widetilde{\boldsymbol{w}}}
        \newcommand{\wo}{\boldsymbol{w}^o}
        \newcommand{\vi}[2]{v_{#1} ({#2})}

        \newcommand{\C}{\boldsymbol{\mathcal{C}}}
        \newcommand{\B}{\boldsymbol{\mathcal{B}}}

        \newcommand{\M}{\boldsymbol{\mathcal{M}}}
        \newcommand{\F}{\boldsymbol{\mathcal{F}}}

        \newcommand{\G}{\boldsymbol{\mathcal{G}}}

        \newcommand{\s}{\boldsymbol{s}}
        
        \newcommand{\BSigma}{\boldsymbol{\Sigma}}
        \newcommand{\bSigma}{\boldsymbol{\Sigma}}
        \newcommand{\bsigma}{\boldsymbol{\sigma}}
        \newcommand{\bgamma}{\boldsymbol{\gamma}}

        \newcommand{\bT}{\boldsymbol{T}}
        
        \newcommand{\bu}{\boldsymbol{u}}
        \newcommand{\bg}{\boldsymbol{g}}
        \newcommand{\I}{\boldsymbol{I}}
        
        \newcommand{\col}{\text{col}}
        \newcommand{\diag}{\text{diag}}

        \newcommand{\cb}[1]{{{\boldsymbol{#1}}}}
        \newcommand{\cp}[1]{\ifmmode {\mathcal{#1}}\else ${\mathcal{#1}}$\fi}
                \newcommand{\cpb}[1]{\ifmmode {{\boldsymbol{\mathcal{#1}}}}\else ${{\boldsymbol{\mathcal{#1}}}}$\fi}

                \newcommand{\bp}{\boldsymbol{p}}
        \newcommand{\bq}{\boldsymbol{q}}
        \newcommand{\bh}{\boldsymbol{h}}
                
        \newcommand{\bs}{\boldsymbol{s}}
        
        \newcommand{\bw}{\boldsymbol{w}}

        \newcommand{\bB}{\boldsymbol{B}}
        \newcommand{\bC}{\boldsymbol{C}}
        \newcommand{\bD}{\boldsymbol{D}}

        \newcommand{\bQ}{\boldsymbol{Q}}
        
        \newcommand{\bP}{\boldsymbol{P}}
        \newcommand{\bH}{\boldsymbol{H}}
        \newcommand{\bA}{\boldsymbol{A}}
        \newcommand{\bR}{\boldsymbol{R}}

        \newcommand{\bZ}{\boldsymbol{Z}}

        \newcommand{\bI}{\boldsymbol{I}}
        \newcommand{\bTheta}{\boldsymbol{\Theta}}

        \newcommand{\Mg}{M_{\nabla}}
                        
        \newcommand{\bvphi}{\boldsymbol{\varphi}}
        
        \newcommand{\tr}{\text{trace}}
        \newcommand{\vc}{\text{vec}}
        
 \usepackage{color}

 \let\oldfrac\frac
 \renewcommand{\frac}[2]{\textstyle{\oldfrac{#1}{#2}}}

\begin{document}
 
     \title{On reducing the communication cost \\ of the diffusion LMS algorithm}

     \author{\IEEEauthorblockN{Ibrahim El Khalil Harrane, R\'emi Flamary, C\'edric Richard, \emph{Senior Member}, \emph{IEEE}}\\
     \IEEEauthorblockA{Universit\'e C\^ote d'Azur, OCA, CNRS, France \\
     ibrahim.harrane@oca.eu, remi.flamary@unice.fr, cedric.richard@unice.fr}
     }

          \maketitle
     ≤
           
 \begin{abstract}
 The rise of digital and mobile communications has recently made the world more connected and networked, resulting in an unprecedented volume of data flowing between sources, data centers, or processes. While these data may be processed in a centralized manner, it is often more suitable to consider distributed strategies such as diffusion as they are scalable and can handle large amounts of data by distributing tasks over networked agents. Although it is relatively simple to implement diffusion strategies over a cluster, it appears to be challenging to deploy them in an ad-hoc network with limited energy budget for communication. In this paper, we introduce a diffusion LMS strategy that significantly reduces communication costs without compromising the performance. Then, we analyze the proposed algorithm in the mean and mean-square sense. Next, we conduct numerical experiments to confirm the theoretical findings. Finally, we perform large scale simulations to test the algorithm efficiency in a scenario where energy is limited. 
 \end{abstract} 
 
 \IEEEpeerreviewmaketitle
     
 \section{Introduction}

 \begin{figure*}[t!]
 \centering
 \includegraphics[height=6cm,keepaspectratio]{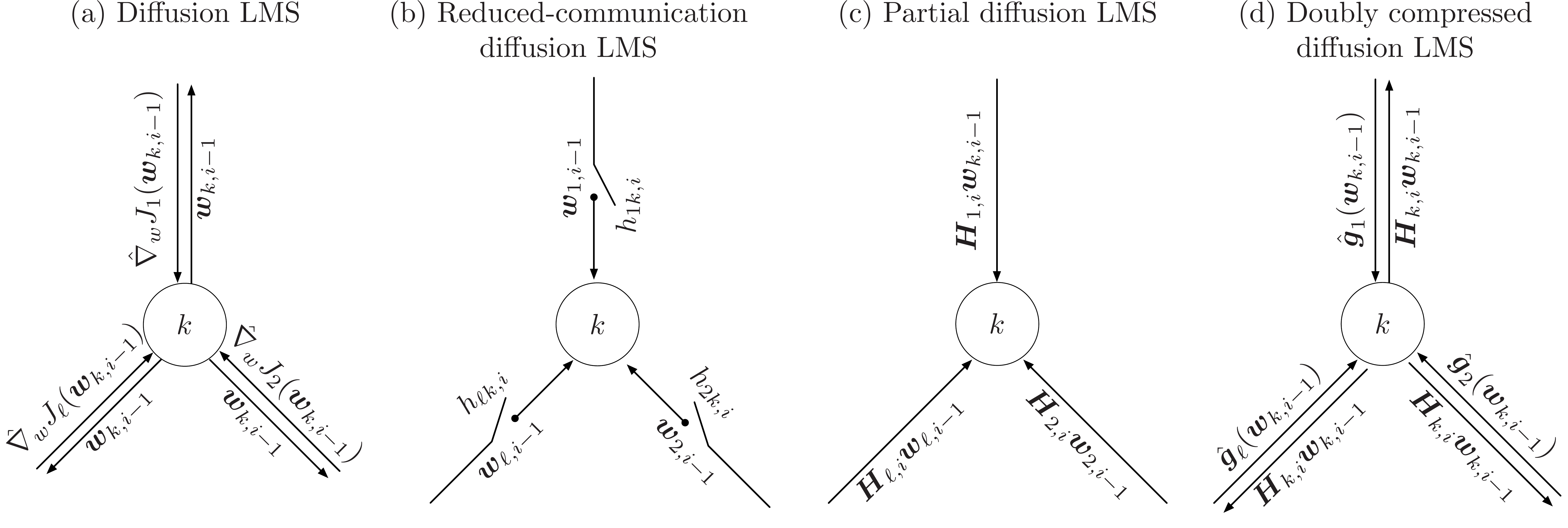}
 \caption{Illustrative representation of transmitted data for the diffusion LMS and different approaches aiming at reducing the communication load for a node $k$.}
 \label{fig:diff_node}
 \end{figure*} 
 
 Adaptive networks are collections of interconnected agents that continuously learn and adapt from streaming measurements to perform a preassigned task such as parameter estimation. The agents are able to share information besides their own data, and collaborate in order to enhance the solution accuracy. Adaptive networks have proven to be powerful tools for modeling natural and social phenomena, ranging from organized organisms to social networks~\cite{sayed2013difadapt}. They are mainly used for data mining tasks over high dimensional data sets locally collected by distributed agents, in a decentralized and cooperative manner. In such scenarios, among other possible strategies~\cite{nedic2009distributed,rabbat2005quantized,lopes2007incremental}, diffusion strategies are a safer option than centralized strategies due to their robustness and resilience to agent and link failures. In particular, the diffusion LMS algorithm plays a central role with its enhanced efficiency and low complexity. It has been studied in single task~\cite{sayed2013difadapt,sayed2013diffusion,sayed2014adaptive} and multitask inference problems~\cite{chen2014diffusion3,chen2014multitask,chen2015diffusion,nassif2015multitask1,nassif2015proximal,chen2016multitask}. Its performance has been analyzed under favorable and unfavorable conditions such as model non-stationarity and imperfect communication ~\cite{takahashi2010diffusion,Khalili2012,Zhao2012impe}. This framework has also been extended by considering more general cost functions and data models~\cite{ChenUCLA2012,ChenUCLA2013,Gharehshiran2013,Chouvardas2011set,sayed2014adaptation}, by incorporating additional regularizers~\cite{Liu2012,Chouvardas2012,Lorenzo2013spar,wen2015diffusion}, or by expanding its use to other scenarios~\cite{Predd2006WSN,chainais2013learning,gao2015diffusion}.

 The advent of the Internet of Things and sensor networks has opened new horizons for diffusion strategies but brought up new challenges as well. Indeed, as illustrated in Fig.\ref{fig:diff_node}~(a), diffusion strategies inherently require all nodes to exchange information with their neighbors at each iteration. In the case of diffusion LMS, as will be detailed in the next section, this information can be either local estimates and gradients of local cost functions, or local estimates only. Even in the latter case, this requirement imposes a substantial burden on communication and energy resources. Reducing the communication cost while maintaining the benefits of cooperation is therefore of major importance for systems with limited energy budget such as wireless sensor networks. In recent years, several strategies were proposed to address this issue. There are mainly two approaches which we illustrate in~Fig.~\ref{fig:diff_node} (b) and (c). On the one hand, some authors propose to restrict the number of active links between neighboring nodes at each time instant~\cite{lopes2008diffusion,arablouei2015analysis}. On the other hand, there are authors that recommend to reduce the communication load by projecting parameter vectors onto lower dimensional spaces before transmission~\cite{sayin2014compressive}, or transmitting only partial parameter vectors~\cite{arablouei2014distributed,arablouei2014adaptive,vadidpour2015partial}. These ideas are related to what is known in the literature as coordinate-descent constructions for single agent optimization. Note that they have recently been extended to distributed settings in~\cite{necoara2017random,xi2017distributed}. They have also been applied to general diffusion networks in~\cite{wang2018coordinate} for general convex cost functions, with a detailed analysis of the performance and stability of the resulting network. In that paper, the authors consider the diffusion LMS and assume that the adaptation step at each node has only access to a random subset of the entries of the approximate gradient vector. At each node, however, all the entries of the local estimates of the neighboring nodes remain available for the combination step. With the exception of some few papers such as~\cite{wang2018coordinate}, the literature mainly focused on the case where the nodes only share a subset of the entries of their local estimates. Nevertheless, it is also of interest to consider the case where both the local estimates and the approximate gradient vectors of the local cost functions are partially shared. 
 This situation may arise due to missing entries. Such schemes are also useful, as considered in this paper, in reducing  communication cost at each iteration in large scale data applications.
        
 In this paper, we propose an algorithm where every transmitted parameter vectors, either local estimates or gradients of local cost functions, are partially shared. The network flow is controlled by two parameters, the number of entries of each one of these two types of parameter vectors.   Then, we study the stochastic behavior of the algorithm in the mean and mean-square sense. Next, we perform numerical experiments to confirm the theoretical findings. Furthermore, we characterize the algorithm performance for high dimensional data in a large network. We compare our algorithm with the diffusion LMS in a sensor network scenario where energy resource is scarce. Finally, we conclude this paper.
 
 \emph{Notation:} Boldface small letters denote vectors. All vectors are column vectors. Boldface capital letters denote matrices. The $(k,\ell)$-th entry of a matrix is denoted by $(\cdot)_{k\ell}$, and the $(k,\ell)$-th block of a block matrix is denoted by $[\,\cdot\,]_{k\ell}$. Matrix trace is denoted by $\text{trace}\{\cdot\}$. The expectation operator is denoted by $\E\{\cdot\}$. The identity matrix of size $N$ is denoted by $\I_{N}$, and the all-one vector of length $N$ is denoted by $\boldsymbol{1}_N$. We denote by $\Nk$ the set of node indices in the neighborhood of node $k$, including $k$ itself, and $|\Nk|$ its cardinality. The operator $\text{col} \{\cdot \}$ stacks its vector arguments on the top of each other to generate a connected vector. The notation $\diag\{a,b\} $ denotes a diagonal matrix with entries $a$ and $b$. Likewise, the notation $\diag\{\boldsymbol{A},\boldsymbol{B}\}$ denotes a block diagonal matrix with block entries $\boldsymbol{A}$ and $\boldsymbol{B}$. The other symbols will be defined in the context where they are used.

 \section{Diffusion LMS and resource-saving variants} 
 
 \subsection{Diffusion LMS}
 
 Consider a connected network composed of $N$ nodes. The aim of each node is to estimate an $L \times 1$ unknown parameter vector $\wo$ from collected measurements. Node $k$ has access to local streaming measurements $\{d_k(i), \bu_{k,i}\} $ where $d_k(i)$ is a scalar zero-mean reference signal, and $\bu_{k,i}$ is an $L\times 1$ zero-mean regression vector with a positive definite  covariance matrix $\bR_{u_k}=\E\{\bu_{k,i}^{\phantom{\top}}\bu_{k,i}^\top\}$. The data at agent $k$ and time $i$ are assumed to be related via the linear regression model:
 \begin{equation}
     \label{eq:datamodel}
     d_k(i) = \bu_{k,i}^\top \bw^{o}+ v_k(i)
 \end{equation}
 where $\bw^o$ is the unknown parameter vector to be estimated, and $v_k(i)$ is a zero-mean i.i.d. noise with variance  $ \sigma_{v,k}^2$.  The noise $v_k(i)$ is assumed to be independent of any other signal. Let $J_k(\bw)$ be a differentiable convex cost function at agent~$k$. In this paper, we shall consider the mean-square-error criterion, namely,
 \begin{equation}
     \label{eq:MSE}
     J_k(\bw) = E \{ |d_k(i) - \bu_{k,i}^\top\bw|^2\}
 \end{equation}
 This criterion is strongly convex, second-order differentiable, and minimized at $\bw^{o}$.
 
 Diffusion LMS strategies seek the minimizer of the aggregate cost function:
 \begin{equation}
     \label{eq:Jglob}
     J^\text{glob}(\bw) = \sum_{k=1}^N J_k(\bw)
 \end{equation}
 in a cooperative manner. Let $\bw_{k,i}$ denote the estimate of the minimizer of \eqref{eq:Jglob} at node~$k$ and time instant $i$. Diffusion LMS algorithm in its Adapt-then-Combine (ATC) form is given by:
 \begin {align}
     &\bpsi_{k,i}	=\w_{k,i-1}-\mu_k \sum_{\ell \in \Nk}c_{\ell k}
     \hat{\nabla}_wJ_\ell(\w_{k,i-1})\label{eq:bpsi}\\
     &\w_{k,i} 		= \sum_{\ell \in \Nk} a_{\ell k}\bpsi_{\ell,i}
     \label{eq:w_diff}
 \end{align}
 with $\hat{\nabla}_wJ_\ell(\w_{k,i-1})=-\bu_{\ell,i}[d_\ell(i) - \bu_{\ell,i}^\top\bw_{k,i-1}]$ the instantaneous approximation of the gradient vector $\nabla_wJ_\ell(\w_{k,i-1})$, $\cp{N}_k$ the neighborhood of node $k$ including $k$, and $\mu_k$ a positive step-size. The nonnegative coefficients $\{a_{\ell k}\}$ and $\{c_{\ell k}\}$ are the $(\ell,k)$-th entries of a left-stochastic matrix $\bA$ and a right-stochastic matrix~$\bC$, respectively.

 \subsection{Reducing the communication load of diffusion LMS}
 
 We shall now describe the existing techniques for reducing the communication load of the diffusion LMS. We start with the reduced communication diffusion LMS (RCD)~\cite{arablouei2015analysis} where each node $k$ can only communicate with a subset of size $m_k$ of its $|\cp{N}_k|$ neighbors. This subset is randomly selected at each node and each iteration. Each agent in the neighborhood of $k$ can be selected with probability:
 \begin{equation}
       p_k= \frac{m_k}{\N_k}
 \end{equation}
 The algorithm can be formulated as:
 \begin{align}
     \begin{dcases}
     \bpsi_{k,i} &= \w_{k,i-1} 
     + \mu_k \bu_{k,i} \left( d_{k,i} - \bu_{k,i}^\top \w_{k,i-1} \right) \\
     \w_{k,i} &=  h_{kk,i}\bpsi_{k,i} 
     + \sum_{\ell \in \Nk\setminus\{k\}} h_{\ell k,i} a_{\ell k} \bpsi_{\ell,i}
     \end{dcases}
     \label{eq:arablouei_2015}
 \end{align}
 with $h_{kk,i}=1-\sum_{\ell \in \Nk\setminus\{k\}}h_{\ell k,i} a_{\ell k}$. Note that matrix $\bC$ in~\eqref{eq:bpsi} has been set to the identity, and $h_{\ell k,i}$ with $\ell\neq k$ is a binary entry depending on whether agent $\ell$ has been selected or not by agent $k$.
     
 Similarly to the RCD, in the distributed LMS with partial diffusion~\cite{arablouei2014distributed,arablouei2014adaptive,vadidpour2015partial}, the matrix $ \bC $ is also set to the identity and the combination step \eqref{eq:w_diff} is now defined as:
 \begin{equation}
     \w_{k,i} =  a_{kk} \bpsi_{k,i} 
     + \!\!\!\!\sum_{\ell \in \Nk\setminus\{k\}}\!\!\!\!
     a_{\ell k} \bigg(\bH_{\ell,i} \bpsi_{\ell,i} 
     +[\bI-\bH_{\ell,i}]\bpsi_{k,i} \bigg) 
     \label{eq:psi_compress2}
 \end{equation}
 where $\bH_{\ell,i}$ is a diagonal entry-selection matrix with $M$ ones and $L-M$ zeros on its diagonal. This means that the nodes can use the entries of their own intermediate estimates in lieu of the ones from the neighbors that have not been communicated. Matrix $\bH_{\ell,i}$ can be deterministic, or can randomly select $M$ entries from all entries. 
 
 Finally, the projection approach investigated in compressive diffusion LMS~\cite{sayin2014compressive} consists of sharing a projection of the local estimates. It also introduces an adaptive correction step to compensate the projection error. This leads to the following formulation of the adaptation step \eqref{eq:w_diff}:
 \begin{align}
     \w_{k,i} &= a_{kk} \bpsi_{k,i} 
     + \sum_{\ell \in \Nk\setminus\{k\}} a_{\ell k} \bgamma_{\ell,i}  
     \label{eq:sayin}
 \end{align}
 where $\bgamma_{\ell,i}= \bgamma_{\ell,i-1} + \eta_\ell \bp_{\ell,i}  \epsilon_{\ell,i} $ is the constructed estimate, with $ \eta_{\ell} $  a positive step-size, $ \bp_{\ell,i} $ a projection vector and $ {\epsilon_{\ell,i}= \bp_{\ell,i}^\top (\bpsi_{\ell,i}-\bgamma_{\ell,i})} $ the reconstruction error. This approach introduces an additional adaptive step which can increase the algorithm complexity.
 
 None of these methods investigates strategies for reducing the communication load induced by the adaptation step \eqref{eq:bpsi} and gradient vector $\hat{\nabla}_wJ_\ell(\cdot)$ sharing. The \emph{doubly compressed diffusion LMS} (DCD) devised in this paper addresses this issue by considering both the adaptation step \eqref{eq:bpsi} and the combination step \eqref{eq:w_diff}.

 \section{Doubly-compressed diffusion LMS}
 
 We shall now introduce our DCD method and analyze its stochastic behavior.  
 The DCD algorithm run at each node $k$ is shown in Alg.~\ref{alg:DCD_LMS}. Matrices $\bH_{k,i}$ and $\bQ_{k,i}$ are diagonal entry-selection matrices with $M$ and $M_\nabla$ ones on their diagonal, respectively. The other diagonal entries of these two matrices are set to zero. First, we consider the adaptation step. The matrix $\bH_{k,i}$ selects $M$ entries (over $L$) of $\bw_{k,i-1}$ that are sent to node~$\ell$ to approximate ${\nabla}_wJ_\ell(\bw_{k,i-1})$ in~\eqref{eq:bpsi}. Node~$\ell$ fills the missing entries of $\bH_{k,i}\bw_{k,i-1}$ by using its own entries $(\I_L - \bH_{k,i}) \w_{\ell,i-1}$, and calculates the instantaneous approximation of the gradient vector at this point. Then node $\ell$ selects $M_\nabla$ entries (over~$L$) of this gradient vector using $\bQ_{k,i}$ and send them to node~$k$. Node~$k$ fills the missing entries by using its own local estimate. Finally, we focus on the combination step. Node $k$ considers the partial parameter vectors $\bH_{\ell,i}\bw_{\ell,i-1}$ received from its neighbors $\ell$ during the adaptation step, and fills the missing entries by using its own local estimate $\bpsi_{k,i}$. Then it aggregate them to obtain the local estimate $\bw_{k,i}$.
 
 \begin{algorithm}[t]
 \caption{Local updates at node $ k $ for DCD }\label{alg:DCD_LMS}
 \begin{algorithmic}[1]
 \Loop
 \State randomly generate  $ \bH_{k,i} $ and $ \bQ_{k,i} $ 
 \For{ $\ell   \in  \cp{N}_k\setminus\{k\} $}
 \State send   $\bH_{k,i} \w_{k,i} $  to node $ \ell $ 
 \State receive from node $\ell$ the partial gradient vector:
 $$\bQ_{\ell,i} \hat{\nabla}_wJ_\ell(\cb{H}_{k,i}\bw_{k,i-1}+(\I_L - \bH_{k,i}) \w_{\ell,i-1})$$
 \State complete the missing entries using those available at node $k$, which results in $\bg_{\ell,i}$ defined in \eqref{eq:g_i} 
 \EndFor
 \State update the intermediate estimate:
 $$\bpsi_{k,i}= \bw_{k,i-1}+\mu_k \sum_{\ell \in \Nk}c_{\ell k} \bg_{\ell,i}$$
 \State calculate the local estimate:
 \begin{align*}
     \w_{k,i} &= a_{kk}\bpsi_{k,i} \nonumber\\ 
     &+\!\!\sum_{\ell \in \Nk\setminus\{k\}}\!\!
     a_{\ell k}\,\left[\bH_{\ell,i}\,\bw_{\ell,i-1} 
     + \left( \I_L - \bH_{\ell,i}  \right) \bpsi_{k,i} \right]
 \end{align*}
 \EndLoop
 \end{algorithmic}
 \end{algorithm}
 
 We can formulate the algorithm in the following form:
 \begin {align}
     \bpsi_{k,i}	&= \bw_{k,i-1}
     +\mu_k \sum_{\ell \in \Nk}c_{\ell k} \bg_{\ell,i} \label{eq:psi_DCD} \\
     \w_{k,i} &= a_{kk}\bpsi_{k,i} \nonumber\\ 
     &+\!\!\sum_{\ell \in \Nk\setminus\{k\}}\!\!
     a_{\ell k}\,\left[\bH_{\ell,i}\,\bw_{\ell,i-1} 
     + \left( \I_L - \bH_{\ell,i}  \right) \bpsi_{k,i} \right] \label{eq:w}
 \end{align}
 with
 \begin{align}
     \label{eq:g_i} 
     &\bg_{\ell,i} \nonumber \\ &=\bQ_{\ell,i} \bu_{\ell,i}\big[d_{\ell}(i)
     -\bu_{\ell,i}^\top (\cb{H}_{k,i}\bw_{k,i-1}
     +(\I_L-\cb{H}_{k,i})\,\bw_{\ell,i-1})\big] \nonumber\\
     &+(\I_L-\cb{Q}_{\ell,i})\bu_{k,i}\big[d_{k}(i) 
     - \bu_{k,i}^\top \bw_{k,i-1})\big]
 \end{align}
 where $\bH_{\ell,i}=\diag \{\bh_{\ell,i}\}$ and $\bQ_{\ell,i}=\diag \{\bq_{\ell,i}\}$. In this paper, we shall assume that $\bh_{\ell,i}$ (resp., $\bq_{\ell,i}$) is an $L \times 1$ binary vector, generated by randomly setting $M$ (resp., $M_\nabla$) of its $L$ entries to $1$, and the other $L-M$ (resp., $L-M_\nabla$) entries to $0$. We shall assume that all possible outcomes for $\bh_{\ell,i}$  (resp., $\bq_{\ell,i}$) are equally likely, and i.i.d. over time and space. Then,
 \begin{equation}
     \E\{\bH_{\ell,i}\}=\frac{M}{L}\I_L 
     \qquad \E\{\bQ_{\ell,i}\}=\frac{M_\nabla}{L}\I_L
     \label{eq:mean_H}
 \end{equation}

 We shall now analyze the stochastic behavior of the DCD algorithm. For the sake of simplicity, we shall consider that matrix $\bC$ is doubly stochastic. We shall also set matrix $\bA$ to the identity matrix. Focusing in this way on the adaptation step and gradient vector sharing helps simplify the analysis. Note that the distributed LMS with partial diffusion~\eqref{eq:psi_compress2}, which exclusively addresses how reducing the communication load induced by the combination step, was analyzed in~\cite{arablouei2014distributed}. Combining both analyses into a single general one is challenging and beyond of the scope of this paper. However, in the sequel, we shall illustrate the efficiency of the DCD algorithm in both cases $\bA=\bI_L$ and $\bA\neq\bI_L$, and compare it with the existing strategies.
 
 Before proceeding with the algorithm analysis, let us introduce the following assumptions on the regression data and selection matrices.
     
     \textbf{Assumption 1} The regression vectors $\uki$ arise from a zero-mean random process that is temporally white and spatially independent. A direct consequence of this assumption is that $\uki$ is independent of $\w_{\ell,j}$ for all $\ell$ and $j<i$.
     
     \textbf{Assumption 2}  The matrices $\bH_{i,k}$ and $  \bQ_{\ell,i} $ arise from a random process that is temporally white, spatially independent, and independent of each other as well as any other process.

 We introduce the $L \times 1$ error vectors:
 \begin{equation}
     \tw_{k,i} = \wo-\w_{k,i} \label{eq:tw_def} 
 \end{equation}
 and we collect them from across all nodes into the vectors:
 \begin{equation}
     \tw_{i} = \col\{ \tw_{1,i}, \tw_{2,i}, \dots, \tw_{N,i} \} \label{eq:twv_def}
 \end{equation}
 Let $\bR_{u_\ell,i}= \bu_{\ell,i}^{\phantom{\top}} \bu_{\ell,i}^\top$. We also introduce:
  \begin{align}
     &\M				=	\diag\{ \mu_1\I_L, \mu_2\I_L, \dots, \mu_N\I_L\} \\	
     &\cb{\cp{R}}_{Q,i}= \diag \bigg\{ \sum_{\ell \in \cp{N}_1}c_{\ell 1}\bQ_{\ell,i} \bR_{u_\ell,i},
             \sum_{\ell \in \cp{N}_2}c_{\ell 2}\bQ_{\ell,i} \bR_{u_\ell,i} ,\dots, \nonumber \\
             & \hspace{3.5cm} 
             \sum_{\ell \in \cp{N}_N}c_{\ell N}\bQ_{\ell,i}   \bR_{u_\ell,i} \bigg\}	\label{eq:R_Q} \\
             &\cb{\cp{H}}_i = \diag \{
                   \bH_{1,i},\bH_{2,i},\dots,\bH_{N,i}\} \\          
           &\cb{\cp{Q}}'_{i} = \diag \bigg\{ \sum_{\ell \in \cp{N}_1}c_{\ell 1}(\I_L-\cb{Q}_{\ell,i}),
         \sum_{\ell \in \cp{N}_2}c_{\ell 1}(\I_L-\cb{Q}_{\ell,i}) \dots, \nonumber \\
             & \hspace{3.5cm} \sum_{\ell \in \cp{N}_N}c_{\ell N}(\I_L-\cb{Q}_{\ell,i}) \bigg\} \label{eq:Qp_def}	 \\ 
             &\cb{\cp{R}}_{u,i} 	= \diag \{ \bR_{u_1,i}, \bR_{u_2,i}, \dots, \bR_{u_N,i} \} \\
     &\cb{\cp{Q}}_i = \diag \{\bQ_{1,i},\bQ_{2,i},\dots,\bQ_{N,i}\} 	\\
     &\C=\BC \otimes \I_L  \label{eq:C_def}
 \end{align}
  where $\otimes$ denotes the Kronecker product. Finally, we introduce the $N \times N $ block matrix  $ \R_{Q(I-H),i} $  with each block $(k,\ell)$ defined as:
 \begin{align}
     [\cb{\cp{R}}_{Q(I-H),i}]_{k \ell}=c_{\ell k} \cb{Q}_{\ell,i}  \bR_{u_\ell,i}  (\I_L-\cb{H}_{k,i})   
     \label{eq:R_Q_I_H}
 \end{align}
   Combining recursion~\eqref{eq:w} and definition~\eqref{eq:tw_def}, and replacing $d_{k}(i)$ by its definition~\eqref{eq:datamodel}, we find:
          \begin{align}
 \tw_{k,i}&=\tw_{k,i-1} \nonumber\\
         &- \mu_k \sum_{\ell \in \Nk}c_{\ell k}
         \bQ_{\ell,i}\bu_{\ell,i}\big[\bu_{\ell,i}^\top\wo+v_{\ell}(i) \nonumber\\
         &\qquad\qquad- \bu_{\ell,i}^\top (\cb{H}_{k,i}\bw_{k,i-1}
         +(\I_L-\cb{H}_{k,i} )\bw_{\ell,i-1})\big] \nonumber \\
         &-\mu_k \sum_{\ell \in \Nk}c_{\ell k}
         (\I_L-\bQ_{\ell,i})\bu_{k,i}\big[\bu_{k,i}^\top\wo
         +v_{k}(i) \nonumber \\ 
         &\qquad\qquad-\bu_{k,i}^\top \bw_{k,i-1}\big] 
 \end{align}
 Note that $\wo = \bH_{k,i}\wo + (\I_L - \bH_{k,i})\wo$. Replacing $ \wo $ by this expression, and using definition~\eqref{eq:tw_def}, leads to:
           \begin{align}
 &\tw_{k,i}=\tw_{k,i-1}
         - \mu_k \sum_{\ell \in \Nk}c_{\ell k} \bQ_{\ell,i}\bu_{\ell,i}\big[\bu_{\ell,i}^\top\cb{H}_{k,i}\tw_{k,i-1} \nonumber \\
         &+\bu_{\ell,i}^\top(\I_L-\cb{H}_{k,i} )\tw_{\ell,i-1} 
         +v_{\ell}(i)\big] \nonumber \\
             &-\mu_k \sum_{\ell \in \Nk}c_{\ell k} (\I_L-\bQ_{\ell,i})\bu_{k,i}\big[\bu_{k,i}^\top \tw_{k,i-1} 
         +v_{k}(i)\big]
         \label{eq:tw_t}
  \end{align}
 Rearranging the terms in \eqref{eq:tw_t}, and using definitions \eqref{eq:twv_def}--\eqref{eq:C_def}, leads to:
         \begin{align}
 \tw_i &=  \big(\I_{NL}
             -\M\cb{\cp{R}}_{Q,i} \cpb{H}_i
             -\M\cb{\cp{Q}}_{i}' \cb{\cp{R}}_{u,i}
             \nonumber \\
             & -\M\cb{\cp{R}}_{Q(I-H),i} \big)\tw_{i-1} 
             -\big( \M\C^\top\cb{\cp{Q}}_i + \M\cb{\cp{Q}}_{i}'\big)\bs_i
             \label{eq:tw_dcmp}	
     \end{align}
 where
 \begin{align}
      \bs_i	=\col\{\bu_{1,i}\vi{1}{i}, \bu_{2,i}\vi{2}{i}, \dots, \bu_{N,i} \vi{N}{i} \} 
 \end{align}

 \subsection{Mean weight behavior analysis}
 
 We shall now examine the convergence in the mean for the DCD algorithm and derive a necessary convergence condition. We start by rewriting the weight-error vector recursion~\eqref{eq:tw_dcmp} as:
 \begin{align}
     \tw_i=\B_i\tw_{i-1}-\cb{\cp{G}}_i \s_i
     \label{eq:tw}
 \end{align}
 where the coefficient matrices $\cpb{B}_i$ and $\cpb{G}_i$ are defined as:
  \begin{align}
     \B_i	&=\I_{NL} \nonumber\\
             &-\M\cb{\cp{R}}_{Q,i} \cpb{H}_i 
              -\M\cb{\cp{Q}}_{i}' \cb{\cp{R}}_{u,i}
              -\M\cb{\cp{R}}_{Q(I-H),i} \label{eq:B_def} \\
             \cb{\cp{G}}_i	&=\M\C^\top\cb{\cp{Q}}_i + \M\cb{\cp{Q}}'_{i} 
 \end{align}
  Taking expectations of both sides of \eqref{eq:tw}, using Assumptions~1 and 2, and $ \E\{\s_i\}=0$, we find:
 \begin{align}
     \E\{\tw_i\}  =  \B \, \E\{\tw_{i-1}\} \nonumber 
 \end{align}
 where
 \begin{align}
 \B &= \I_{NL} 
         -\frac{M \Mg}{L^2}\M\cb{\cp{R}} 
             - \left(  1 - \frac{\Mg}{L} \right ) \M\cb{\cp{R}}_u   \label{eq:E_B_i}  \\
             & \hspace{12mm} - \frac{\Mg}{L} \left(1-\frac{M}{L} \right)\M\C^\top\!\cb{\cp{R}}_{u} \nonumber \\
     \cb{\cp{R}}_{u}  &= \E \{ \cb{\cp{R}}_{u,i}Â \} = \diag \{ \bR_{u_1}, \bR_{u_2}, \dots, \bR_{u_N} \} \\
     \R 			         &= \diag \{ \bR_{1},\ldots,\bR_{N} \}
     \end{align}
     with 
     \begin{align}
         \bR_{k} = \sum_{\ell  \in \cp{N}_k} c_{\ell ,k} \bR_{u_\ell}
     \end{align}
             
 From \eqref{eq:E_B_i}, we observe that the algorithm~\eqref{eq:w} asymptotically converges in the mean toward $\wo$ if, and only if, 
 \begin{equation}
     \label{eq:cond-stab}
     \rho ( \B )<1
 \end{equation}
 where $\rho(\cdot)$ denotes the spectral radius of its matrix argument. We know that $\rho(\boldsymbol{X})\leq\|\boldsymbol{X}\|$ for any induced norm. Then:
         \begin{align}
             \rho ( \B ) \leq & \| \B   \|_{b,\infty}  \nonumber \\
      \leq & \max_{k, \ell} \| [ \B ]_{k \ell} \|
      \label{eq:spectral.radius}
         \end{align}
  where $\|\!\cdot\!\|_{b,\infty}$ denotes the block maximum norm.  From \eqref{eq:spectral.radius} we have:
 \begin{equation}
 \label{eq:spectral.radius2}
 \begin{split}
    &\rho ( \B ) \leq   
     \max_{\ell,k} \Big\| \I_{L}
     -\mu_k\Big[\frac{M \Mg}{L^2} \bR_k
     + \frac{M}{L}\left(1-\frac{\Mg}{L} \right )\bR_{u_k}\\
     &\hspace{2cm} + \frac{\Mg}{L} \left(1-\frac{M}{L} \right)
     c_{\ell k}  \bR_{u_\ell}\Big]\Big\|
 \end{split}    
 \end{equation}
 As a linear combination with positive coefficients of positive definite matrices $\bR_k$ and $ \bR_{u_\ell} $, the matrix in square brackets on the RHS of \eqref{eq:spectral.radius2} is positive definite. Condition~\eqref{eq:cond-stab} then holds if:
 \begin{equation}
     \mu_{k} <  \oldfrac{2}{ \lambda_{\max,k}}
 \end{equation}
 for all $k$, with 
 \begin{equation}
     \begin{split}
 \lambda_{\max,k} &= \frac{M \Mg}{L^2}  \lambda_{\max}( \bR_k) 
     +  \frac{M}{L} \left(  1 - \frac{\Mg}{L} \right ) \lambda_{\max}( \bR_{u_k} ) \\
     & +  \frac{\Mg}{L} \left(1-\frac{M}{L} \right)  \max\limits_{\ell \in \cpb{N}_k} \,  \, c_{\ell k}  \, \lambda_{\max} (\bR_{u_\ell} )
     \end{split}
 \end{equation}
 where we used Jensen's inequality with $\lambda_{\max}(\cdot)$ operator, that stands for the maximum eigenvalue of its matrix argument.
 It is worth mentioning that when $ M = \Mg = L $ we retrieve the convergence condition of the diffusion LMS as derived in \cite{sayed2013difadapt}:
 \begin{equation}
 \lambda_{\max,k} = \lambda_{\max}( \bR_k) 
 \end{equation}
           With this setting, we also retrieve the matrices $\cpb{B}_i$ and $\cpb{G}$ of the diffusion LMS in~\cite[(262)--(263)]{sayed2013difadapt}.
 \color{black}
 
 \subsection{Mean-square error behavior analysis}
 
 We are now interested in providing a global solution for studying the mean square error. With this aim, we consider the weighted square measure $\E\{\|\tw_{i}\|^2_{\bSigma}\}$ where $\bSigma$ denotes a $N \times N$ block diagonal weighting matrix. By setting $\bSigma$ to different values, we can extract various types of information about the nodes and the network such as the network mean square deviation MSD, or the excess mean square error EMSE.
 
 We start by using the independence Assumption~1 and~\eqref{eq:tw} to write
 \begin{equation}
     \label{eq:tw_2}
             \E \|\tw_i\|^2_{\BSigma}=\E\{\tw_{i-1}^\top\B_i^\top\BSigma\B_i\tw_{i-1}\}
     + \E\{\s_i^\top\G_i^\top\BSigma\,\G_i\s_i\}
 \end{equation}
 On the one hand, the second term on the RHS of~\eqref{eq:tw_2} can be written as:
     \begin{align}
         \E\{\s_i^\top \G_i^\top\BSigma\,\G_i\s_i\} 
         &=\tr\big(\E\{\s_i^\top \G_i^\top\BSigma\,\G_i \s_i\}\big)						\nonumber \\
         &=\tr\big(\E\{\G_i ^\top\BSigma\,\G_i \}\, \E\{\s_i\s_i^\top\}\big)					\nonumber \\
         &=\tr\big(\E\{\G_i ^\top\BSigma\,\G_i \}\, \cpb{S}\big)							 
     \label{eq:traceSigR}
     \end{align}
 where
 \begin{equation} 
     \cpb{S}=\E \{ \s_i \s_i^\top \} = \diag(\sigma_{v,1}^2\cb{R}_{u,1},\ldots,\sigma_{v,N}^2\cb{R}_{u,N}) 
 \end{equation}
 and
         \begin{align}
         &\E\{\G_i ^\top\BSigma\,\G_i \} \nonumber \\ & =  \big( \M \C^\top \cb{\cp{Q}}_{i} + \M \cb{\cp{Q}}_{i}' \big)^\top \bSigma 
                             \big( \M \C^\top \cb{\cp{Q}}_{i} + \M \cb{\cp{Q}}_{i}' \big) \nonumber \\
           &=\bTheta_1 + \bTheta_2 +\bTheta_2^\top +\bTheta_3
         \label{eq:y}
          \end{align}
      with
      \begin{align}
     & \bTheta_1= \E \{ \cb{\cp{Q}}_i \C \M \bSigma \M \C^\top \cb{\cp{Q}}_i\}  \label{eq:Theta_1} \\
     & \bTheta_2= \E \{ \cb{\cp{Q}}_i \C \M \bSigma \M \cb{\cp{Q}}_{i}'\}   \label{eq:Theta_2} \\
     & \bTheta_3= \E \{ \cb{\cp{Q}}_{i}' \M \bSigma \M  \cb{\cp{Q}}_{i}'\}   \label{eq:Theta_3}
      \end{align}
 
 Before proceeding with the calculation of \eqref{eq:Theta_1}--\eqref{eq:Theta_3}, we introduce some preliminary results.
 
 Given any $L \times L$ matrix $\BSigma$, it can be shown:
 \begin{align}
        &\E\{\bQ_{\ell,i} \BSigma \bQ_{k,i} \}= \label{eq:cov_Q_ll} \\
        &\begin{dcases} 
        \frac{M_\nabla}{L} 
        \left( \left(1- \frac{M_\nabla-1}{L-1}\right) \I_{L} \odot \bSigma 
        + \frac{M_\nabla-1}{L-1} \bSigma \right) & \text {if  } \ell =k \\
        \left(\frac{M_\nabla}{L}\right)^2 \BSigma & \text {otherwise}
        \end{dcases} \nonumber
 \end{align}
 where $\odot$ is the Hadamard entry-wise product. 
 
 Consider the block diagonal matrix ${\cb{\cp{Q}}}_i$ and any $NL \times NL$ matrix $\cb{\Pi}$. By using \eqref{eq:cov_Q_ll} for each block $\E\{[\cb{\cp{Q}}_{i}\cb{\Pi}\cb{\cp{Q}}_{i}]_{k\ell}\}$, it follows that:
 \begin{equation}
     \label{eq:EQ}
     \begin{split}
     &\E\{\cb{\cp{Q}}_{i}\cb{\Pi}\cb{\cp{Q}}_{i}\}  \\  
     &= \alpha_1  \, ( \I_{N} \otimes \boldsymbol{1}_{LL}) \odot \cb{\Pi} + \alpha_2  \, \I_{NL} \odot \cb{\Pi} + \alpha_3  \, \cb{\Pi}
     \end{split}
 \end{equation}
 
 where $\boldsymbol{1}_{LL}$ denotes the all-one $L \times L$ matrix, and:
 \begin{align}
     \alpha_1  &= \frac {\Mg}{L} 
     \left(\frac{\Mg - 1}{L- 1} - \frac{\Mg}{L} \right) \\
     \alpha_2  &= \frac {\Mg}{L} \left(  1 - \frac{\Mg - 1}{L- 1}  \right) \\  
     \alpha_3  &=  \left(   \frac{\Mg }{L}  \right)^2 
     \label{eq:alpha_def}
 \end{align}
 
 Finally we consider the $NL \times NL$ matrix, say $\cb{\varphi}_Q(\cb{\Pi})$, defined by its $L \times L$ blocks:
 \begin{equation}
     \label{eq:PhiQ-block}
     [\cb{\varphi}_Q(\cb{\Pi})]_{k \ell}=
     \E\{\bQ_{k,i} [\cb{\Pi} ]_{k \ell} \bQ_{k,i} \}
 \end{equation}
 By using~\eqref{eq:cov_Q_ll} for each block, it can be shown that $\cb{\varphi}_Q(\cb{\Pi})$ can be expressed as follows:
 \begin{equation}
     \label{eq:PhiQ}
     \cb{\varphi}_Q(\cb{\Pi})
     =\alpha_2   \, (\boldsymbol{1}_{NN} \otimes \I_{L}) \odot \cb{\Pi}  
     +(\alpha_1  + \alpha_3 ) \, \cb{\Pi}
 \end{equation}
 Note that $\cb{\varphi}_Q(\cb{\Pi})=\E\{\cb{\cp{Q}}_{i}\cb{\Pi}\cb{\cp{Q}}_{i}\}$ if $\cb{\Pi}$ is block diagonal.
 
 We can now proceed with the evaluation of \eqref{eq:Theta_1}--\eqref{eq:Theta_3}.  Matrix $\bTheta_1$ calculation follows by setting $\cb{\Pi}=\C \M \bSigma \M \C^\top$ in~\eqref{eq:PhiQ}.
 Consider now $\bTheta_2$ in~\eqref{eq:y}. We have:
 \begin{align}
 [\bTheta_2]_{k \ell}  &=  
 \frac{\Mg}{L}  [\C \M \BSigma \M]_{k \ell} 
 -  c_{k \ell }^2 \, \E \{ \bQ_{k,i}  [ \M \BSigma \M]_{\ell \ell} \bQ_{k ,i}  \}  \nonumber \\
 & -   \left( \frac{\Mg}{L} \right)^2  \left(  [\C \M \BSigma \M]_{k \ell} 
 -  c_{k \ell }^2 \,  [ \M \BSigma \M]_{\ell \ell}  \right )
 \end{align}
 We can use~\eqref{eq:PhiQ-block} to calculate the second term in the RHS of the above equation since $\M\BSigma\M$ is block diagonal. This yields:
 \begin{equation}
     \label{eq:y2_t}
     \begin{split}
     \bTheta_2  &=  \frac{\Mg}{L}  \C \M \BSigma \M  
               -   \C_2 \cb{\varphi}_Q(\M\BSigma\M)  \\
             & -   \left( \frac{\Mg}{L} \right)^2  \left(  \C \M \BSigma \M
              -     \C_2 \M \BSigma \M\right )  
          \end{split}
 \end{equation}
 where $\C_2 = \C \odot \C $. 
      
 Finally, we calculate the last term $\bTheta_3$ in the RHS of~\eqref{eq:y}. Matrix $\bTheta_3$ is block diagonal, with each diagonal block defined as follows:
 \begin{equation*}
     \begin{split}
        &[\bTheta_3]_{k k} \\  
         &=\E\{[\cb{\cp{Q}}_{i}']_{kk}[\M\bSigma\M]_{k k}
         [\cb{\cp{Q}}_{i}']_{kk}\} \\
         &=\sum_{m,n=1}^N c_{mk} c_{n k}\,\E\{\left(\I_L-\bQ_{m,i}\right)
         [\M\bSigma\M ]_{k k}\left(\I_L-\bQ_{n,i}\right)\}
     \end{split}
 \end{equation*}
 Using~\eqref{eq:mean_H}, we get:
      \begin{align}
         &[ \bTheta_3]_{k k} = \left( 1 - 2 \frac {\Mg}{L} \right)  [\M \bSigma \M ]_{k k} \sum_{m,n=1}^N c_{mk} c_{n k}  \nonumber  \\
              &\hspace{1cm} + \sum_{m,n=1}^N c_{mk} c_{n k}   \E \{ \bQ_{m,i}  [\M \bSigma \M ]_{k k} \bQ_{n,i}   \}  \nonumber 
     \end{align}
 Applying~\eqref{eq:cov_Q_ll} leads to:
      \begin{align}
         &[ \bTheta_3]_{k k} = \left( 1 - 2 \frac {\Mg}{L} \right)  [\M \bSigma \M ]_{k k} \nonumber \\
              & \hspace{1cm} + \sum_{m}^N c_{mk}^2 \E \{ \bQ_{m,i}  [\M \bSigma \M ]_{k k} \bQ_{m,i}   \} \nonumber \\
             & \hspace{1cm} +  \left(  \frac{\Mg}{L} \right)^2 \left ( \sum_{m , n=1}^N c_{mk} c_{n k}    [\M \bSigma \M ]_{k k} \right.  \nonumber  \\
             & \hspace{1cm}  - \left.  \sum_{m =1}^N c_{mk}^2     [\M \bSigma \M ]_{k k}   \right) \label{eq:y3_t}
         \end{align}
 Finally, using~\eqref{eq:PhiQ}, we can write~\eqref{eq:y3_t} in a compact form:
         \begin{align}
          \bTheta_3  &= \left( 1 - 2 \frac {\Mg}{L} \right)  \M \bSigma \M  
              +    ( \I_{NL} \odot  \C\, \C^\top  ) \, \bvphi_Q( \M \bSigma \M )  \nonumber \\
             &+   \left(  \frac{\Mg}{L} \right)^2 \left (  \M \bSigma \M  
             -     ( \I_{NL} \odot  \C\, \C^\top  )  \M \bSigma \M    \right)
          \end{align}
   
   It is interesting to notice that the second term in the RHS of \eqref{eq:tw_2} does not depend on $M$. Moreover, setting $\Mg=L$ results in canceling $ \bTheta_2$ and $\bTheta_3$ since $\cb{\cp{Q}}'_{i}=\cb{0}$.

 On the other hand, the first term on the RHS of \eqref{eq:tw_2} depends on both parameters $M$ and $\Mg$ and can be expressed as:
 \begin{equation}
     \E \{ \tw_{i-1}^\top\B_i^\top\BSigma\B_i\tw_{i-1} \}=\E\|\tw_{i-1}\|_{\BSigma'} ^2
             \label{eq:w_Sig_w}
 \end{equation}
 where the weighting matrix $\bSigma'$ is defined as:
     \begin{equation}
             \BSigma'= \E \{ \B_i^\top \BSigma \B_i \}
         \label{eq:Sigma'_def}
         \end{equation} 
 Replacing $ \cpb{B}_i$ by its definition \eqref{eq:B_def} leads to:
     \begin{equation}
     \begin{split}
         \BSigma'	=\BSigma
             &-  \frac{M \Mg}{L^2}  \BSigma \M\cb{\cp{R}}
             - \left(  1 - \frac{\Mg}{L} \right ) \BSigma \M\cb{\cp{R}}_u  \nonumber \\
             &- \frac{\Mg}{L} \left(1-\frac{M}{L} \right)   \BSigma \M\C^\top\cb{\cp{R}}_{u}  \nonumber \\
             &- \frac{M \Mg}{L^2}  \cb{\cp{R}}  \M  \BSigma  
          - \left(  1 - \frac{\Mg}{L} \right ) \cb{\cp{R}}_u \M \BSigma      \nonumber \\ 
         & - \frac{\Mg}{L} \left(1-\frac{M}{L} \right)   \cb{\cp{R}}_{u} \C^\top \M \BSigma\nonumber \\
         &+ \sum_{j=1}^{6} \bP_j 
         + \bP_2^\top +\bP_{3}^\top + \bP_{5}^\top 	
         \end{split}
 \end{equation}
 where
     \begin{align}
     &\bP_1= \E \{ \cpb{H}_i \cb{\cp{R}}_{Q,i}^\top\M \BSigma \M \cb{\cp{R}}_{Q,i} \cpb{H}_i \} \\
     &\bP_2= \E \{ \cpb{H}_i \cb{\cp{R}}_{Q,i}^\top\M \BSigma \M \cpb{Q}_{i}' \cb{\cp{R}}_{u,i} \} \\
     &\bP_3= \E \{ \cpb{H}_i \cb{\cp{R}}_{Q,i}^\top\M \BSigma \M  \cb{\cp{R}}_{Q(I-H),i}  \} \\
     &\bP_4= \E \{ \cb{\cp{R}}_{u,i}^\top \cpb{Q}_{i}' \M \BSigma \M \cpb{Q}_{i}' \cb{\cp{R}}_{u,i} \} \\
     &\bP_5= \E \{  \cb{\cp{R}}_{u,i}^\top \cpb{Q}_{i}' \M \BSigma \M  \cb{\cp{R}}_{Q(I-H),i}  \} \\
     &\bP_6= \E \{ \cb{\cp{R}}_{Q(I-H),i}^\top  \M \BSigma \M  \cb{\cp{R}}_{Q(I-H),i} \} 
     \end{align}
 
 Due to the complexity for calculating the terms $\bP_{j}$ and the outcomes, we shall not report them in this section. Instead, we provide all the necessary steps and  results in the Appendix.
 
 Following the same reasoning as in~\cite{sayed2013difadapt}, we express $\bSigma'$ in a vector form as:
 \begin{equation}
     \bsigma'=\F \bsigma
 \end{equation}
 where 
 \begin{align*}
 &\bsigma = \vc(\bSigma)  
 &\bsigma' = \vc(\bSigma')
 \end{align*}
 and the coefficient matrix  $ \cpb{F} $  of size $ (MN)^2 \times (MN)^2 $ is defined as:
 \begin{align}
     \F &= \I_{(NM)^2}
             -  \frac{M \Mg}{L^2}  \cb{\cp{R}} \M \otimes \I_{LN} \nonumber\\
             &-  \left(  1 - \frac{\Mg}{L} \right ) \cb{\cp{R}}_u  \M \otimes \I_{LN}\nonumber \\
         & - \frac{\Mg}{L} \left(1-\frac{M}{L} \right)  \cb{\cp{R}}_{u} \C \M \otimes \I_{LN} 
           - \frac{M \Mg}{L^2} \, \I_{LN} \otimes \cb{\cp{R}}  \M  \nonumber \\
         &- \left(  1 - \frac{\Mg}{L} \right )  \, \I_{LN} \otimes \cb{\cp{R}}_u \M \nonumber \\
         &- \frac{\Mg}{L} \left(1-\frac{M}{L} \right)    \, \I_{LN} \otimes \cb{\cp{R}}_{u} \C^\top \M \nonumber \\
         & +\sum_{j=1}^6 \bZ_j  
              +\bZ_{2^\top} +\bZ_{3^\top}  + \bZ_{5^\top }
     \end{align}  
     \normalsize
 where the matrices $\bZ_j$ and $ \bZ_{j^\top} $ are obtained when applying the $\vc(\cdot)$ operator to $\bP_j$ and $ \bP_j^\top $, respectively,  as it is shown in the Appendix.
  
 Substituting \eqref{eq:traceSigR} and \eqref{eq:w_Sig_w} into~\eqref{eq:tw_2}, and applying the $\vc(\cdot)$ operator to both sides, we get:
 \begin{align}
     \E\|\tw_i\|^2_{\bsigma}=\E \|\tw_{i-1}\|^2_{\F\bsigma}
     +\tr\big(\E \{\G^\top \BSigma\,\G \}\,\cpb{S}\big)
     \label{eq:E_w_2_2}
 \end{align}
 
 Using \eqref{eq:E_w_2_2}, it is possible to extract useful information about the network or a specific node. For instance, we calculate the network mean square deviation or excess mean square error by setting $\bSigma=\I_{LN}$ and $\bSigma=\R_u$, respectively. The DCD can be seen as an extension of the diffusion LMS in the case where the weighting matrix $\bA$ is the identity matrix. Indeed, it is possible to recover the diffusion LMS, and derive other variants such as the compressed diffusion LMS, by properly setting matrices $\{\bH_{k,i},\bQ_{k,i}\}$ and parameters $\{M,\Mg\}$.
 
 \section{Simulation Results}
 \label{sec:simulations}
 
 In this section, we shall first evaluate the accuracy of the mean-square error behavior model. Then, we shall perform two experiments to characterize the performance of the DCD algorithm compared to the diffusion LMS algorithm, the reduced-communication diffusion LMS~\cite{arablouei2015analysis}, and the partial diffusion LMS \cite{arablouei2014adaptive}. We shall also consider the so-called compressed diffusion LMS (CD) obtained by setting $\bA=\bI_L$ and $\bQ_{\ell,i}=\bI_L$ in \eqref{eq:psi_DCD}--\eqref{eq:w}, which means that $\Mg=L$ in this case. Before proceeding, note that the compression ratios of the CD and DCD algorithms are equal to $\frac{2L}{M+L}$ and $\frac{2L}{M+\Mg}$.
 
 First we considered a small network to validate the theoretical model. Then, we used a larger network and high dimensional measurements with the aim of testing the algorithm in a larger scale setting. Finally, we considered an energy dependent network where the agents alternate between active and inactive states depending on the available energy.  For the three experiments, the parameter vectors $\wo$ were generated from a zero-mean Gaussian distribution. The input data $\bu_{k,i}$ were drawn from zero-mean Gaussian distributions, with covariance $\bR_{u,k}= \sigma_{u,k}^2 \I_L$ reported in Fig.~\ref{fig:model_validation} (right).
 The weighting matrices $\bC$ were generated using the Metropolis rule~\cite{sayed2013difadapt}. Noises $v_k(i)$ were zero-mean, i.i.d. and Gaussian distributed with variance  $\sigma_{v,k}^2=10^{-3}$. Simulation results were averaged over $100$ Monte-Carlo runs.

 \subsubsection{Experiment 1}
       \begin{figure}[t]
     \centering
     \includegraphics[width=.22\textwidth,height=4.5cm,keepaspectratio]{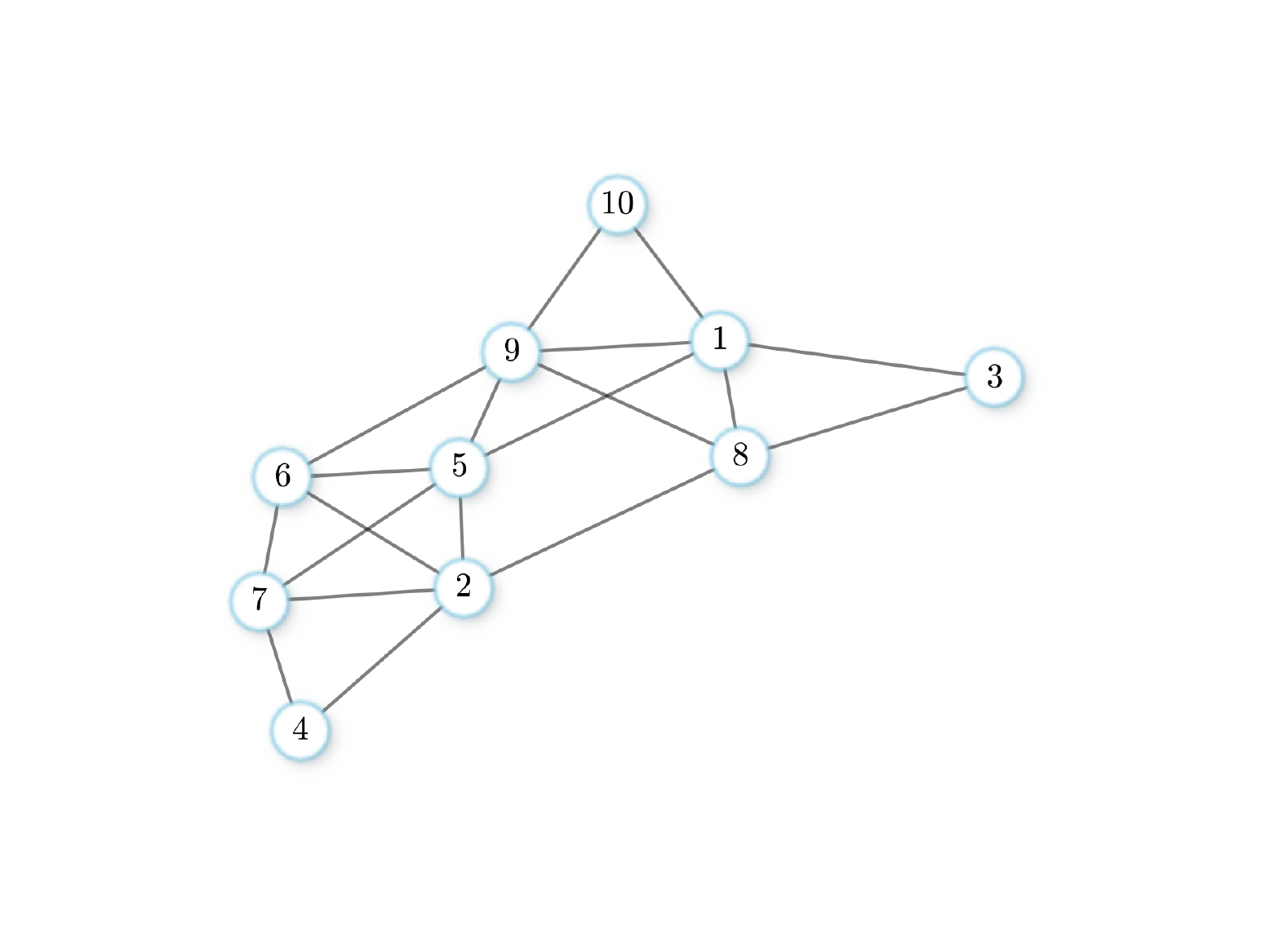}
     \includegraphics[width=.22\textwidth,height=4cm, keepaspectratio]{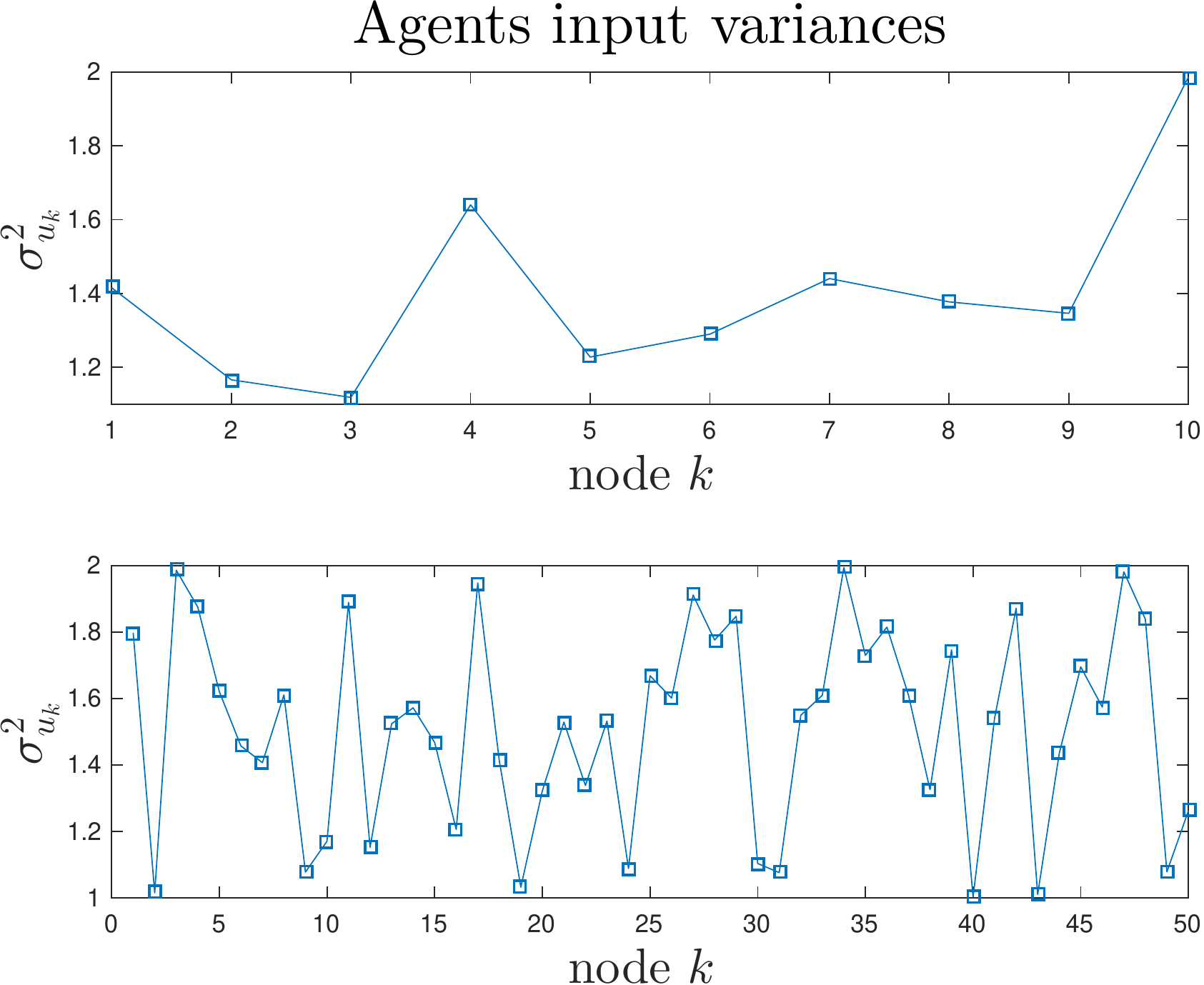}
     \caption{ (left) Network topology. (right) Variance $\sigma^2_{u_k}$ of regressors in Experiment 1 (top) and Experiment 2 (Bottom). }
     \label{fig:model_validation}
 \end{figure}
\begin{figure*}[t]
     \centering
     \includegraphics[width=.32\textwidth, height=5 cm, keepaspectratio]{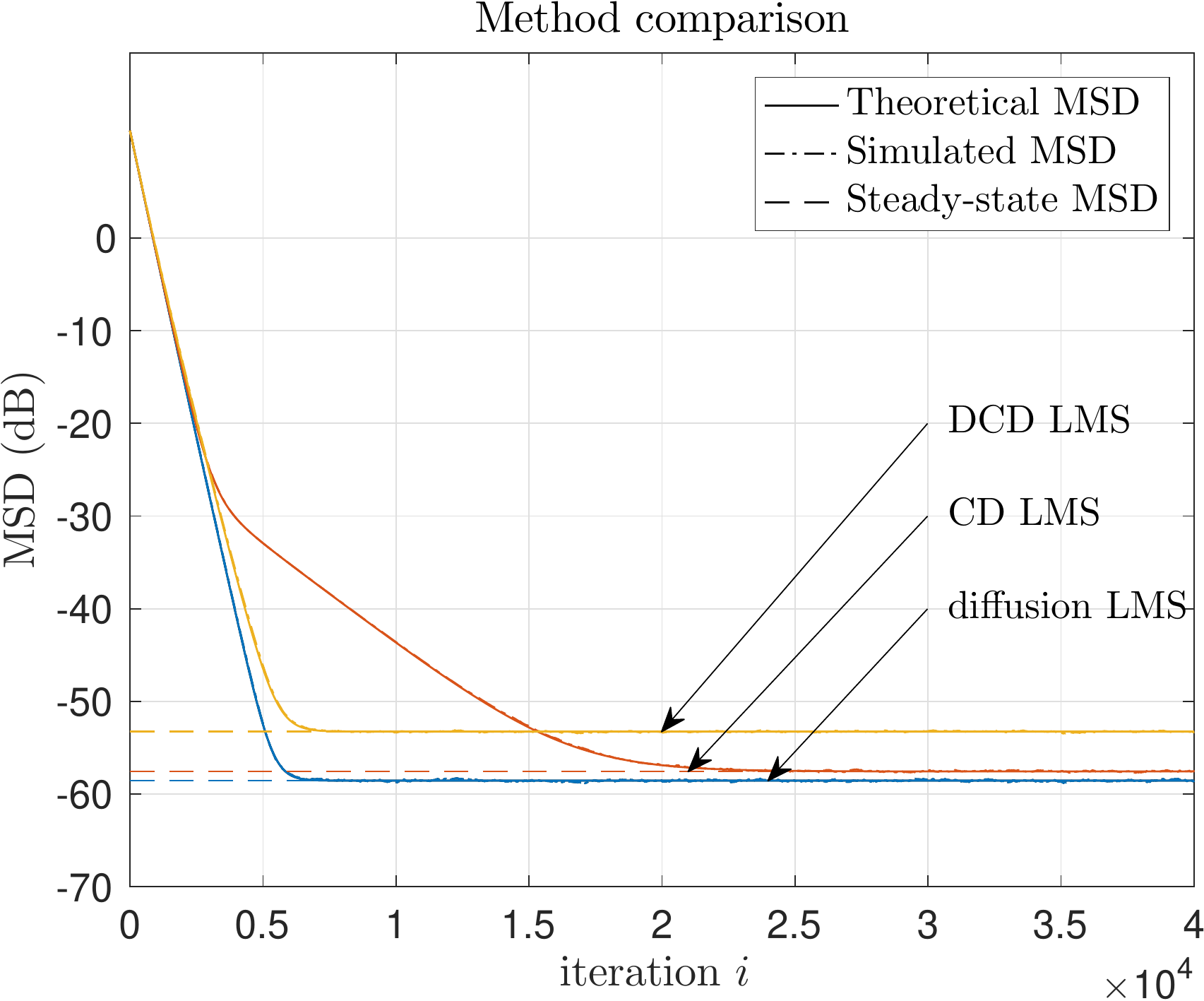}
     \includegraphics[width=.32\linewidth , height=5 cm,keepaspectratio ]{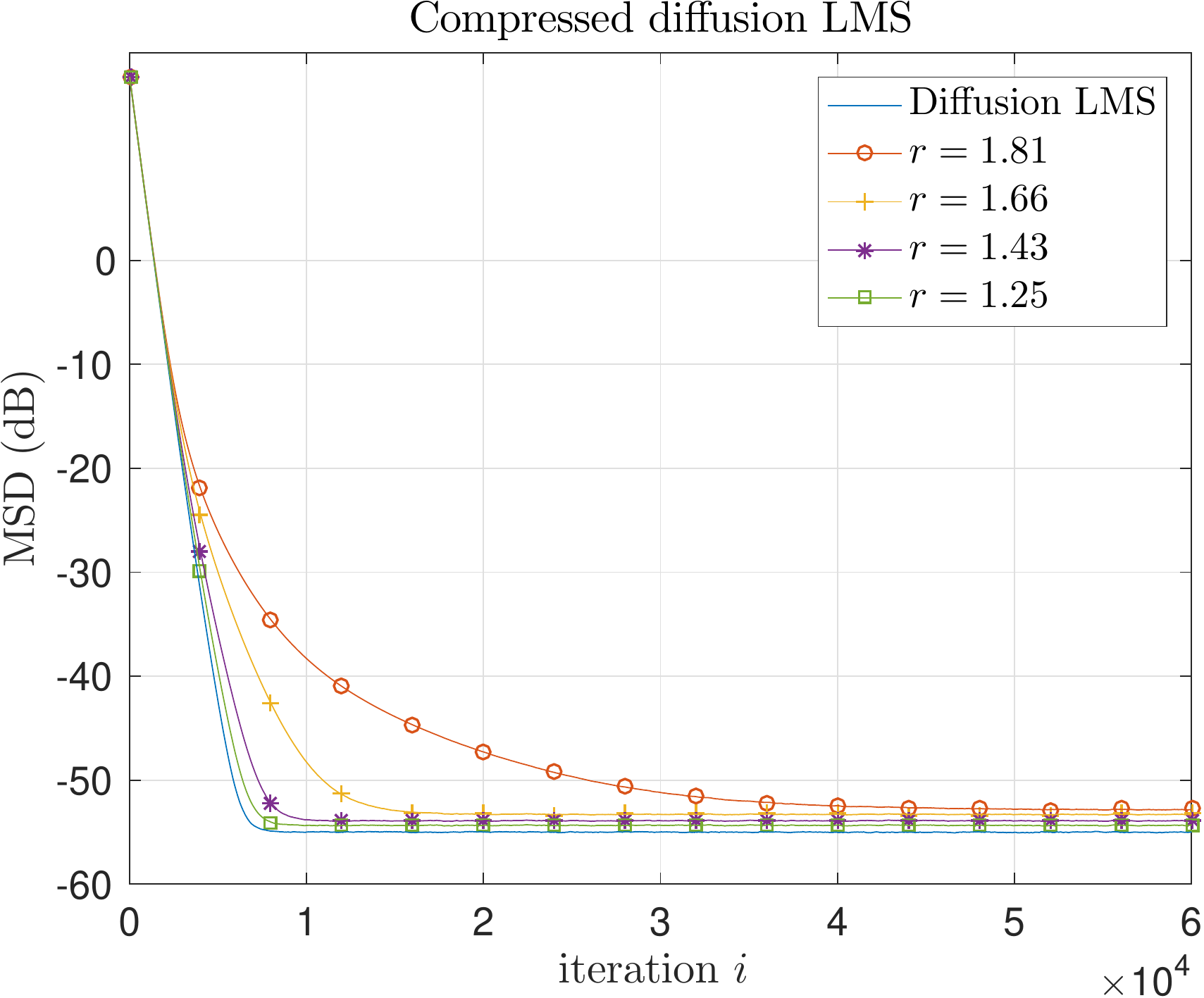}
     \includegraphics[width=.32\linewidth , height=5 cm,keepaspectratio ]{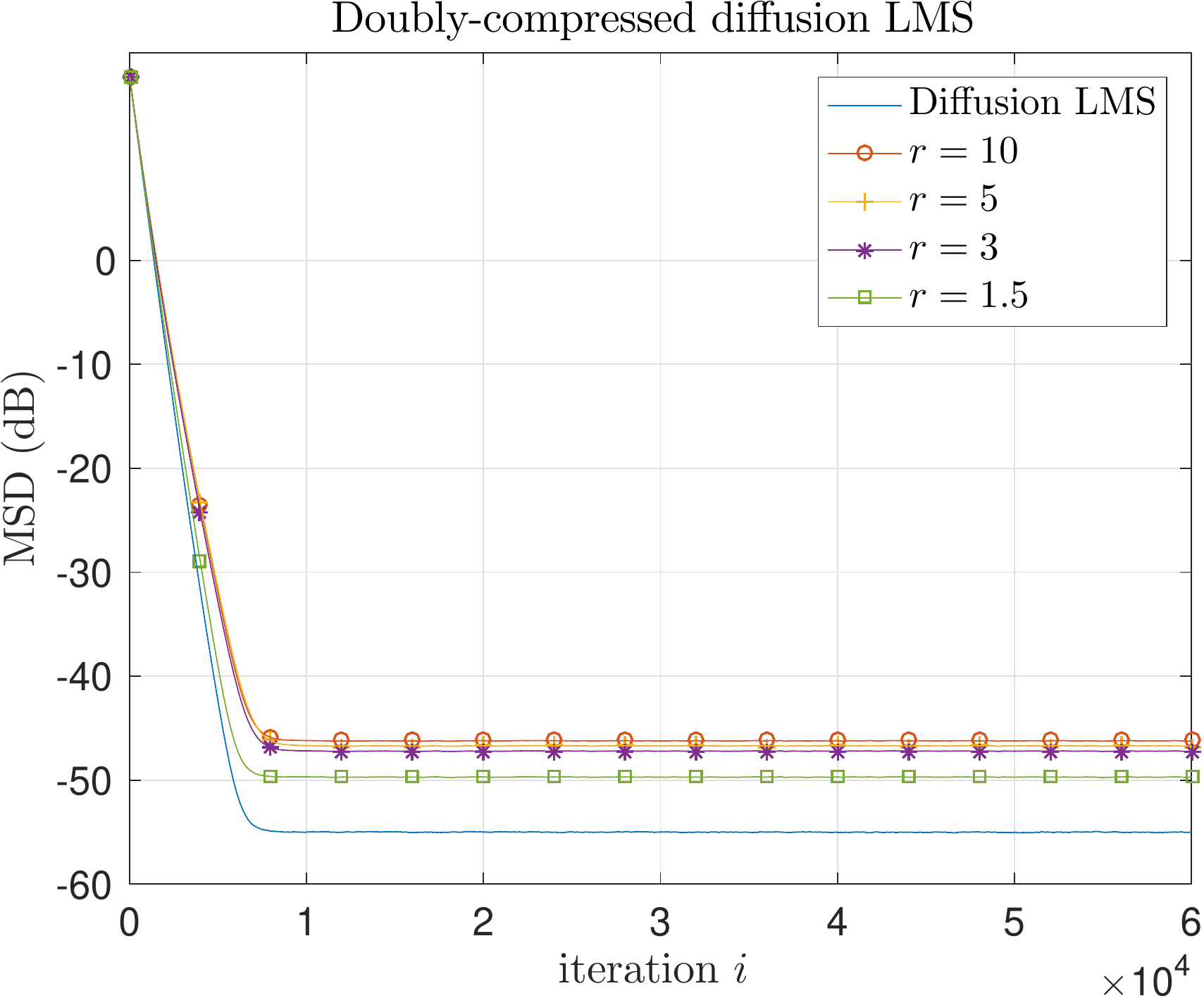}
     \caption{ (left) Theoretical and simulated MSD curves for diffusion LMS and its compressed versions.  Evolution of the MSD as a function of the compression ratio for {compressed diffusion LMS} (center),  and {doubly-compressed diffusion LMS} (right).}
     \label{fig:simus}
 \end{figure*}

   
 We considered the network with $N=10$ nodes depicted in Fig.~\ref{fig:model_validation} (left). We set the parameters as follows: $\mu_k=10^{-3}$, $L=5$, $M=3$, $M_\nabla=1$. This resulted in compression ratio of $\frac{10}{8}$ and $\frac{10}{4}$ for compressed diffusion and doubly compressed diffusion LMS, respectively. It can be observed in Fig.~\ref{fig:simus} (left) that the theoretical model accurately fits the simulated results. Unsurprisingly, the diffusion LMS algorithm outperformed its compressed counterparts at the expense of a higher communication load.
 
 \subsubsection{Experiment 2}
 Since compression is particularly relevant for relatively large data flows, then we  considered a network with $N=50$ agents. We set the algorithm parameters as follows: $\mu_k=3 \cdot 10^{-2}$, $L=50$, $M=5$.
  Due to the high dimensionality of the matrix $\F $ ($ 2500^2 \times 2500^2 $), we only performed  Monte-Carlo simulations using C language scripts. Figure~\ref{fig:simus} depicts the performance of the algorithms for different compression ratios. The largest compression ratio that can be reached by the CD algorithm equals $\frac{100}{55}$ as it transmits the whole gradient vectors ($\bQ_{\ell,i}=\bI_L$). On the other hand, the CDC diffusion~LMS offers more flexibility and can adapt to the network communication load by adjusting $M$ and $\Mg$.

 \subsubsection{Experiment 3}
 
 In a realistic wireless sensor network (WSN) implementation, nodes have limited energy reserves and cannot be active all the time. One of the most promising solution for this issue is to adopt an ENO strategy, where ENO stands for Energy Neutral Operation. In other words, the agents consume at most the amount of energy they harvest, hence achieving the neutral energy condition. Theoretically, neutral energy condition guarantees an infinite sensor lifetime. 
 In order to implement an ENO strategy, nodes must be endowed with energy harvesting and storage capabilities. Agents alternate between two phases: a brief active phase and a sleeping phase. During the active phase, each agent $k$ performs its assigned tasks and calculates the duration $T_{s_k,i}$ of the sleeping phase based on the available energy, the consumed energy and an estimate of the energy that will be harvested~\cite{le2013multi}. For the sake of limiting energy consumption, the agents then switch to sleep mode for a duration of $T_{s_k,i}$. The corresponding DCD based algorithm is presented in Alg.~\ref{alg:DCD_LMS_en}.
 
 \begin{algorithm}[t]
 \caption{Local updates at node $k$ for the modified DCD }\label{alg:DCD_LMS_en}
 \begin{algorithmic}[1]
 \Loop
 \State randomly generate  $ \bH_{k,i} $ and $ \bQ_{k,i} $ 
 \For{ $\ell   \in  \cp{N}_k\setminus\{k\} $}
 \State send   $\bH_{k,i} \w_{k,i} $  to node $ \ell $ 
 \State receive from node $\ell$ the partial gradient vector:
 $$\bQ_{\ell,i} \hat{\nabla}_wJ_\ell(\cb{H}_{k,i}\bw_{k,i-1}+(\I_L - \bH_{k,i}) \w_{\ell,i-1})$$
 \State complete the missing entries using those available at node $k$, which results in $\bg_{\ell,i}$ defined in \eqref{eq:g_i} 
 \EndFor
 \State update the intermediate estimate:
 $$\bpsi_{k,i}= \bw_{k,i-1}+\mu_k \sum_{\ell \in \Nk}c_{\ell k} \bg_{\ell,i}$$
 \State calculate the local estimate:
 \begin{align*}
     \w_{k,i} &= a_{kk}\bpsi_{k,i} \nonumber\\ 
     &+\!\!\sum_{\ell \in \Nk\setminus\{k\}}\!\!
     a_{\ell k}\,\left[\bH_{\ell,i}\,\bw_{\ell,i-1} 
     + \left( \I_L - \bH_{\ell,i}  \right) \bpsi_{k,i} \right]
 \end{align*}
 \State switch and stay in sleep mode for $T_{s_k,i}$ seconds
 \EndLoop
 \end{algorithmic}
 \end{algorithm}
 
 We considered a solar energy based WSN with Bluetooth capabilities. To calculate $T_{s_k,i}$, we used~\cite{le2013multi}:
 \begin{equation}
     \displaystyle
     T_{s_k,i}
     =\frac{e_{c_k,i}-\eta\,e_{s_k,i} }{\eta\,(P_{\text{harv},k,i}-P_\text{leak})-P_\text{sleep}}
 \end{equation}
 where $e_{c_k,i}$ and $e_{s_k,i}$ denote the consumed energy and the stored energy, respectively, $\eta$ is the power manager efficiency, $P_{\text{harv},k,i}$ is the harvested power, $P_\text{leak}$ is the capacitor leakage power, and $P_\text{sleep}$ is the power consumed during sleep phase. These parameters and  other parameters used for the experiment are defined in Table~\ref{tab:T_s_parmeters}.
 
 \begin{table}[t]
 \center
 \caption{Summary of the parameters used to determine the duration of sleeping phase $ T_{s} $  }
 \label{tab:T_s_parmeters}
 \begin{tabular}{| c | c | c |}
 \hline
 parameter & description & value \\
 \hline
 $ C_s $ & super capacitor capacity & $0.09 $ F	 \\
 \hline
 $P_\text{leak}$ & super capacitor leakage power  &  $ 3.3 \cdot  10^{-6}$ W \\
 \hline
 $P_\text{sleep}$ & consumed power for sleep mode & $ 3.01 \cdot 10^{-5}$ W \\
 \hline
 $T_{s_{\min}}$ &  minimal sleep time duration & $ 1 $ s \\
 \hline
 $T_{s_{\max}}$ & maximal sleep time duration & $ 300 $ s  \\
 \hline
 $V_\text{ref}$ & minimal required voltage  & $ 3.5 $ V \\
 \hline
 $ e_{a,\text{diff}} $	& consumed energy for diffusion LMS &  $ 8.58\cdot 10^{-2}$ J \\
 \hline
 $ e_{a,\text{RCD}} $	& consumed energy for red. comm. LMS &  $1.61\cdot 10^{-2}$ J \\
 \hline
 $ e_{a,\text{PM}} $	& consumed energy for part. dif. LMS &  $5.4\cdot 10^{-3}$ J \\
 \hline
 $ e_{a,\text{cmp}} $	& consumed energy for CD LMS &  $7.51\cdot 10^{-2}$ J \\
 \hline
 $ e_{a,\text{dcmp}} $	& consumed energy for DCD LMS &  $  5.4\cdot 10^{-3} $ J \\
 \hline
 \end{tabular}
 \end{table}
 \begin{figure*}[t]
     \centering
     \includegraphics[width=.33\textwidth,height=6.cm, keepaspectratio]{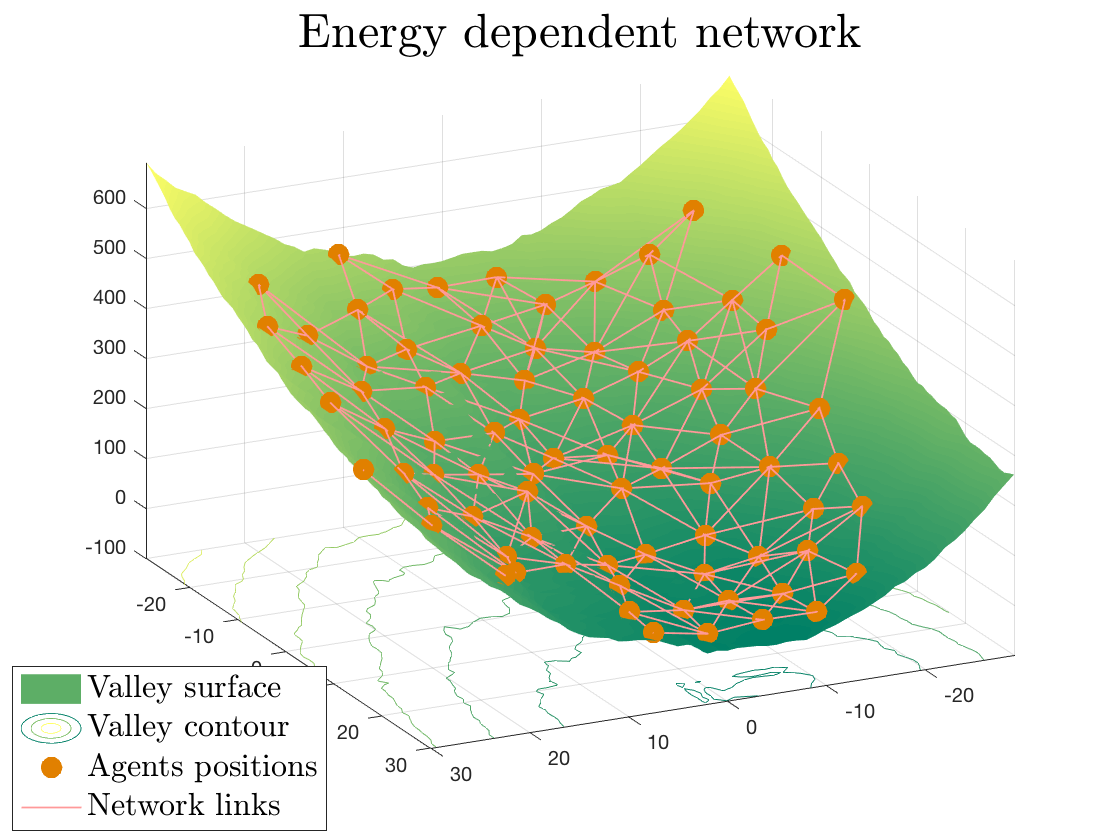}
     \includegraphics[width=.3\linewidth , height=6cm,keepaspectratio ]{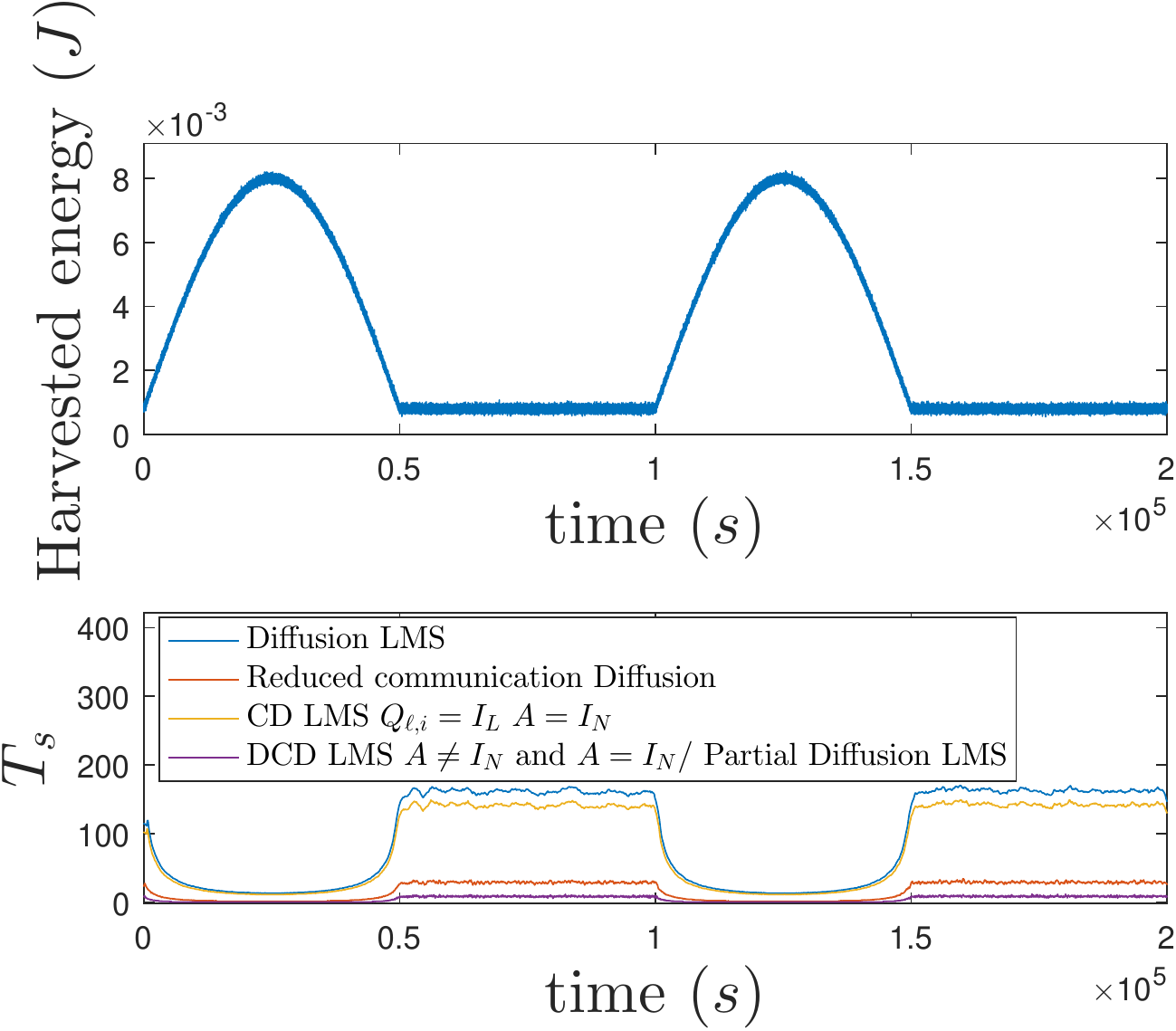}
     \includegraphics[width=.31\linewidth , height=7cm,keepaspectratio ]{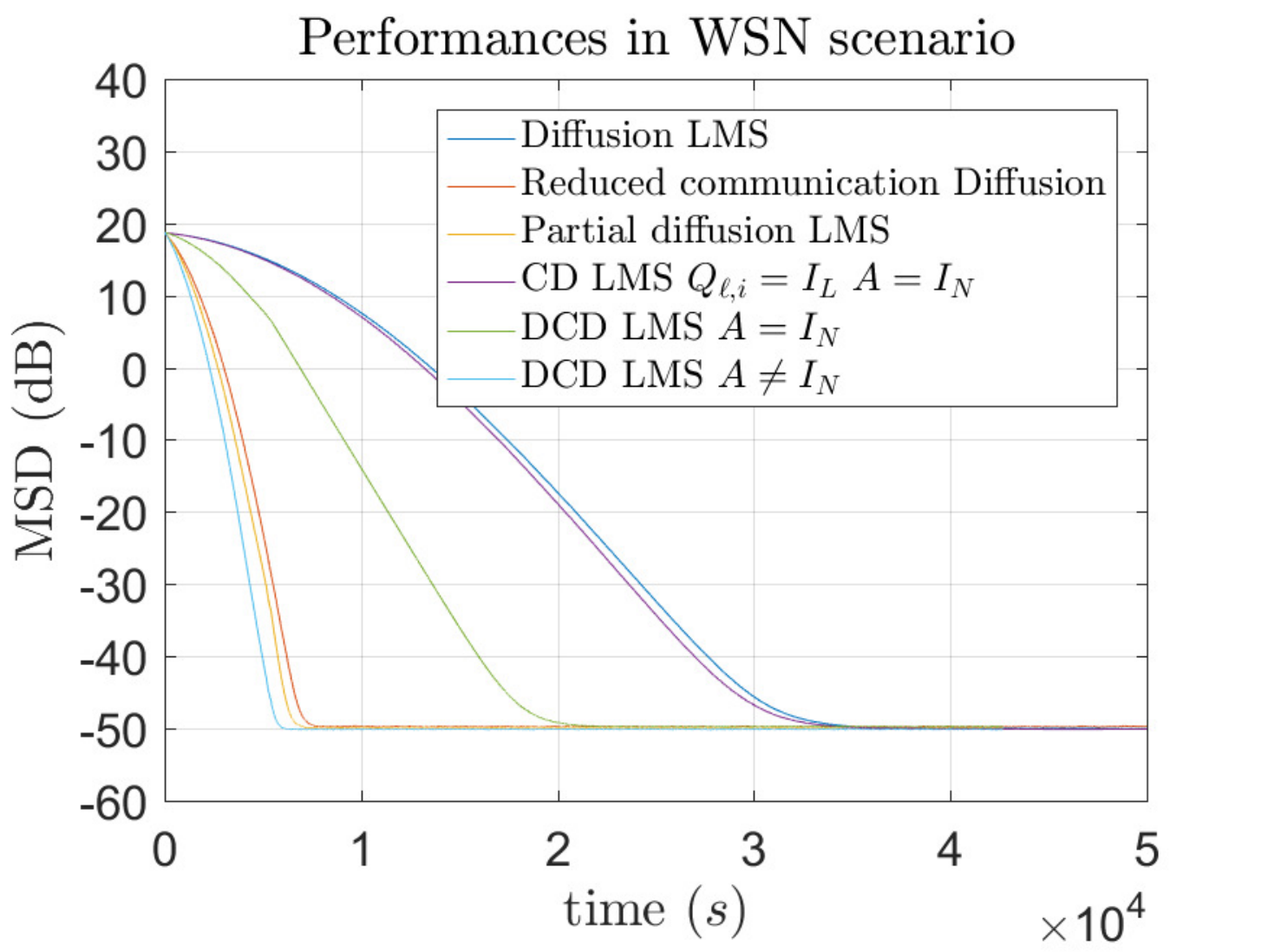}
     \caption{ (left) Network topology for WSN experiment. (center) Harvested energy and sleep periods during the experimentations. (right)  Simulated  MSD curves.}
     \label{fig:ENO_fig}
 \end{figure*}
 Following~\cite{le2013multi}, we estimated $e_{c_k,i}$ as follows:
 \begin{equation}
     e_{c_k,i} = e_{a} + P_\text{sleep} T_{{s_k},{i-1}}
     \label{eq:ec}
 \end{equation}
 where $e_a$ denotes the consumed energy during the active phase, assumed to be constant and known, and $P_\text{sleep} T_{{s_k},{i-1}}$ is a prediction of the consumed energy during the sleep phase $i$ based on the duration of the sleep phase $i-1$. Quantity $e_a$ depends on the algorithm. It is essentially dictated by the volume of transfered data because of the excessive energy consumption of the Bluetooth module. As $P_\text{sleep}$, it was determined based on our own measurements and an estimation of the number of frames sent by each algorithm. See Table~\ref{tab:T_s_parmeters}.
 
 Finally, we considered the following law to simulate the amount of harvested energy:
 \begin{equation}
     E_{\text{harv},k,i} = \max(0, E_0 \sin(2 \pi f i) + n(i))
 \end{equation}
 with $E_0 = 0.67 J$, $E_{\text{harv},k,i}$ the harvested energy at node~$k$ and time $i$, $f=10^{-5}$ a frequency, and $n_k(i) $ a zero-mean 
 Gaussian noise with variance $\sigma_n^2=10^{-6} $. Note that the additive noise was used to diversify the amount of harvested energy during the Monte-Carlo runs. While it would have been possible to use a constant value over time for the harvested energy, we induced periodicity through the $\sin(\cdot)$ function to roughly model solar energy. We considered the network in Fig. \ref{fig:ENO_fig} (left). It consists of $N=80$ agents scattered over a hill with different lighting levels. We set $L$ to $40$.  To compare the algorithms, we set their compression ratio to $r=20$. One exception was made for the CD algorithm. As parameter $r$ cannot reach such a large value, it was set to $r=\frac{80}{65}$. Next the step size of each algorithm was set according to Table~\ref{tab:step_sizes} in order to reach the same steady-state MSD. When $\bA\neq\bI_L$, matrix $\bA$ was set using the Metropolis rule~\cite{sayed2013difadapt}.

 \begin{table}[t]
 \center
 \caption{Step-size and compression settings for the different tested algorithms.}
 \label{tab:step_sizes}
 {\renewcommand\arraystretch{1.25}
 \begin{tabular}{| c | c | c | }
 \hline
 Algorithm & Step-size $ \mu_k $  & Comp. ratio \\
 \hline
 Diffusion LMS & $ 5.4 \, 10^{-3} $  & / \\
 \hline
 Reduced communication diffusion \cite{arablouei2015analysis} &  $1.14 \cdot \, 10^{-2} $ & $ 20 $ \\
 \hline
 Partial diffusion LMS \cite{arablouei2014adaptive} &  $ 4.4 \cdot 10^{-3} $  & $ 20 $ \\
 \hline
 Compressed diffusion LMS & $ 4.8 \cdot 10^{-2} $  & $ \frac{80}{65} $ \\
 \hline
 Doubly-compressed diffusion LMS &  $ 6 \cdot 10^{-3} $  & $ 20 $\\
 \hline
 \end{tabular}}
 \end{table}
 
 We shall now discuss the results in Fig.~\ref{fig:ENO_fig}. Figure~\ref{fig:ENO_fig} (center) shows that the sleep phase duration decreases as the amount of harvested energy increases, and conversely. Also note that, for all the algorithms, the sleep phase is longer at the beginning as a consequence of the limited amount of stored energy that is available. Next, the sleep phase duration drops down until it reaches the minimal sleep duration $T_{s_{\min}}$ if possible. The less energy an algorithm consumes, the faster the super capacitors charge, and the faster the sleep phase duration of the agents drop down. As a consequence, nodes can process larger amounts of data, which makes the convergence of the algorithm faster as confirmed in Fig.~\ref{fig:ENO_fig} (right). This can be observed with the diffusion LMS and the DC algorithm, which are outperformed by the other algorithms. Let us now focus on the partial diffusion LMS and the DCD algorithm $(\bA\neq\bI_L)$. As their compression ratio $r$ was set to same value for comparison purposes, and their consumed energy during the active phase is almost the same, their sleep phases in Fig.~\ref{fig:ENO_fig} (center) are superimposed. The DCD algorithm however outperformed the partial diffusion LMS, in particular because it is endowed with a gradient sharing mechanism. Both algorithms outperformed the reduced-communication diffusion LMS.

 \section{Conclusion}
 
 Among the challenges brought up by the advent of the Internet of Things and WSN, energy efficiency is a critical one. To address this challenge, we investigated a technique for diffusion LMS that consists of sharing partial data. We carried out an analysis of the stochastic behavior of the proposed algorithm in the mean and mean-square sense. Furthermore, we provided simulation results to illustrate the accuracy of the theoretical models. Finally, we considered a realistic simulation where sensor nodes alternate between active and inactive phases. This experiment confirmed the efficiency of the proposed strategy.

 \section{Appendix} 
 
 Before proceeding with the calculation of the terms $\bP_j$, we introduce some preliminary results.
 
 Given any $L \times L$ matrix $\BSigma$, it can be shown:
 \begin{align}
        &\E\{\bH_{\ell,i} \BSigma \bH_{k,i} \}= \label{eq:cov_H_ll} \\
        &\begin{dcases} 
        \frac{M}{L} 
        \left( \left(1- \frac{M-1}{L-1}\right) \I_{L} \odot \bSigma 
        + \frac{M-1}{L-1} \bSigma \right) & \text {if  } \ell =k \\
        \left(\frac{M}{L}\right)^2 \BSigma & \text {otherwise}
        \end{dcases} \nonumber
 \end{align}
 where $\odot$ is the Hadamard entry-wise product. 
 
 Consider the block diagonal matrix ${\cb{\cp{H}}}_i$ and any $NL \times NL$ matrix $\cb{\Pi}$. By using \eqref{eq:cov_H_ll} for each block $\E\{[\cb{\cp{H}}_{i}\cb{\Pi}\cb{\cp{H}}_{i}]_{k\ell}\}$, it follows that:
 \begin{equation}
     \label{eq:EQ_1}
     \begin{split}
     &\E\{\cb{\cp{H}}_{i}\cb{\Pi}\cb{\cp{H}}_{i}\}  \\  
     &= \beta_1  \, ( \I_{N} \otimes \boldsymbol{1}_{LL}) \odot \cb{\Pi} + \beta_2  \, \I_{NL} \odot \cb{\Pi} + \beta_3  \, \cb{\Pi}
     \end{split}
 \end{equation}
 
 where $\boldsymbol{1}_{LL}$ denotes the all-one $L \times L$ matrix, and:
 \begin{align}
     \beta_1  &= \frac {M}{L} 
     \left(\frac{M - 1}{L- 1} - \frac{M}{L} \right) \\
     \beta_2  &= \frac {M}{L} \left(  1 - \frac{M - 1}{L- 1}  \right) \\  
     \beta_3  &=  \left(   \frac{ M }{L}  \right)^2 
     \label{eq:beta_def}
 \end{align}
 
 Finally we consider the $NL \times NL$ matrix, say $\cb{\varphi}_H(\cb{\Pi})$, defined by its $L \times L$ blocks:
 \begin{equation}
     \label{eq:PhiH-block}
     [\cb{\varphi}_H(\cb{\Pi})]_{k \ell}=
     \E\{\bH_{k,i} [\cb{\Pi} ]_{k \ell} \bH_{k,i} \}
 \end{equation}
 By using~\eqref{eq:cov_H_ll} for each block, it can be shown that:
 \begin{equation}
     \label{eq:PhiH}
     \cb{\varphi}_H(\cb{\Pi})
     =\beta_2   \, (\boldsymbol{1}_{NN} \otimes \I_{L}) \odot \cb{\Pi}  
     +(\beta_1  + \beta_3 ) \, \cb{\Pi}
 \end{equation}
 Note that $\cb{\varphi}_H(\cb{\Pi})=\E\{\cb{\cp{H}}_{i}\cb{\Pi}\cb{\cp{H}}_{i}\}$ if $\cb{\Pi}$ is block diagonal.
 
 \subsection{Terms $\bP_j$ calculation}
 
 \subsubsection{Term $\bP_1$ calculation}
 
 Matrix $\bP_1$ is a block diagonal matrix. Its $k$-th diagonal block is given by:
 \begin{equation}
     [\bP_1]_{k k}  
     = \E \{ \bH_{k,i} [\cb{\cp{R}}_{Q,i}]_{k k}^\top [ \M \BSigma \M ]_{k k} 
     [ \cb{\cp{R}}_{Q,i} ]_{k k} \bH_{k,i} \} 
 \end{equation}
 \normalsize
 Substituting $\cb{\cp{R}}_{Q,i}$ by its expression~\eqref{eq:R_Q} leads to:
 \begin{align}
     [\bP_1]_{k k}  &= \sum_{m,n=1}^N c_{mk} c_{nk} \\ &\quad\E \{ \bH_{k,i} \bR_{u_m,i}
     \bQ_{m,i} [ \M \BSigma \M ]_{k k} 
     \bQ_{n,i} \bR_{u_n,i}  \bH_{k,i} \} \nonumber
 \end{align}
 We rewrite $[\bP_1]_{k k}$ as a sum of two terms, one for $m = n$ and one for $m \neq n$. Using~\eqref{eq:mean_H}, we get:
 \begin{align}
      &[\bP_1]_{k k}  = \nonumber\\ 
      & \sum_{m=1}^N c_{mk}^2 \E \{ \bH_{k,i} \bR_{u_m,i} 
      \bQ_{m,i} [ \M \BSigma \M ]_{k k} \bQ_{m,i} \bR_{u_m,i}\bH_{k,i}\}\nonumber \\
      &+   \Big(\frac{\Mg}{L}\Big)^2  \Big(\nonumber\\ 
      &\phantom{+}\sum_{m,n=1}^N c_{mk} c_{nk} \E \{ \bH_{k,i}\bR_{u_m,i}
      [\M\BSigma\M ]_{k k}\bR_{u_n,i}\bH_{k,i}\}- \nonumber \\
      &\phantom{+}\sum_{m=1}^N c_{mk}^2 \E \{ \bH_{k,i}\bR_{u_m,i}
      [\M\BSigma\M ]_{k k} \bR_{u_m,i}  \bH_{k,i}\}\Big) 
     \label{eq:P_1_t_1}
     \end{align}
     The terms in~\eqref{eq:P_1_t_1} depends of higher-order moments of the regression data. While we can continue the analysis by calculating these terms, it is sufficient for the exposition to focus on the case of sufficiently small step-sizes where a reasonable approximation is~\cite{sayed2013difadapt}:  
 \begin{equation}
     \E \{ \bR_{u_m,i}  [ \M \BSigma \M ]_{k k} \bR_{u_m,i} \} = \bR_{u_m}  [ \M \BSigma \M ]_{k k} \bR_{u_m}
 \end{equation}
 Note that this approximation will also be used in the sequel.
 
           Finally, using Assumption 2, we can reformulate $\bP_1 $ as:
     \begin{align}
       &\bP_1  = \sum_{m=1}^N \E \{ \cpb{H}_i \cpb{R}_{c_m} \bvphi_Q( \M \BSigma \M ) \cpb{R}_{c_m} \cpb{H}_i \} \label{eq:P1} \\
      &+\Big(\frac{\Mg}{L} \Big)^2  \Big(\nonumber\\ 
      &\E \{ \cpb{H}_i \cpb{R}\M\BSigma\M\cpb{R} \cpb{H}_i \} 
      - \sum_{m=1}^N\E\{\cpb{H}_i\cpb{R}_{c_m}\M\BSigma\M\cpb{R}_{c_m}\}\cpb{H}_i \Big)  \nonumber 
     \end{align}
 \normalsize
 where the matrices $\cpb{R}_{c_k} $ are defined as:
     \begin{equation}
             \label{eq:R_m_def}
             \cpb{R}_{c_k}  = \diag \{  c_{k 1} \bR_{u_k} , \dots ,c_{k N} \bR_{u_k} \}  
      \end{equation}
                                              \normalsize
 \subsubsection{Term $\bP_2$ calculation }
 
  Using the same steps as above, we have:
     \begin{align}
      [\bP_2]_{k k} =  \E \{ \bH_{k,i} [\cb{\cp{R}}_{Q,i}]_{k k}^\top [ \M \BSigma \M ]_{k k} [ \cpb{Q}_{i}' ]_{k k} \bR_{u_k,i} \} \nonumber 
      \end{align}
 \normalsize
 We substitute $  [\cb{\cp{R}}_{Q,i}]_{k k} $ and $ [ \cpb{Q}_{i}' ]_{k k} $ by their respective definitions \eqref{eq:R_Q} and \eqref{eq:Qp_def}:
      \begin{align}
    & [\bP_2]_{k k} = \sum_{m,n=1}^N c_{mk} c_{nk}  \E \{\bH_{k,i}  \bR_{u_m,i} \bQ_{m,i} [ \M \BSigma \M ]_{k k} \nonumber  \\ 
         & \hspace{3cm} \left( \I_L - \bQ_{n,i} \right)  \bR_{u_k,i}  \}   \nonumber 
          \end{align}
 \normalsize
 Using \eqref{eq:mean_H} we find that:
      \begin{align}
     & [\bP_2]_{k k}  =  \frac{M \Mg}{L^2} \sum_{m=1}^N c_{mk}^2  \E \{ \bR_{u_m,i}  [ \M \BSigma \M ]_{k k}   \bR_{u_k,i}   \} \nonumber \\
     & \hspace{2mm}-  \frac{M}{L} \sum_{m=1}^N c_{mk}^2  \E \{ \bR_{u_m,i} \bQ_{m,i} [ \M \BSigma \M ]_{k k}  \bQ_{m,i}   \bR_{u_k,i}   \}    \nonumber \\
     & \hspace{2mm}+  \frac{M \Mg}{L^2}  \left( 1 - \frac{\Mg}{L}  \right) \nonumber \\
     & \hspace{1.2cm}\Big( \sum_{m,n=1}^N c_{mk} c_{nk}  \E \{  \bR_{u_m,i} [ \M \BSigma \M ]_{k k}   \bR_{u_k,i}   \} \nonumber \\
     &\hspace{1.2cm} -  \sum_{m=1}^N c_{mk}^2  \E \{ \bR_{u_m,i}  [ \M \BSigma \M ]_{k k}   \bR_{u_k,i}   \}   \Big)
      \end{align}
 \normalsize
 Finally, we write:
      \begin{align}
     & \bP_2   \\&=  \frac{\Mg M}{L^2}  \cpb{R}_{2}   \M \BSigma \M    \cpb{R}_{u} 
      - \frac{ M}{L}  \cpb{R}_{2}  \bvphi_Q( \M \BSigma \M)\cpb{R}_{u}\nonumber  \\
      & + \frac{M \Mg}{L^2}  \left( 1 - \frac{\Mg}{L}  \right) \left(  \cpb{R}   \M \BSigma \M    \cpb{R}_{u}  \nonumber
     - \cpb{R}_{2}   \M \BSigma \M    \cpb{R}_{u}     \right)
     \end{align}
 \normalsize
 where
     \begin{equation}
     \cpb{R}_2 =\left \{  \sum_{m=1}^N c_{m 1}^2 \bR_{u_m} , \dots , \sum_{m=1}^N c_{m N}^2 \bR_{u_m} \right \} 
     \end{equation}
 
 \subsubsection{ Term $\bP_3$ calculation}
 
 Term $\bP_3$ can be expressed as
      \begin{align}
     [\bP_3]_{k \ell} =  \E \{ \bH_{k,i} [\cb{\cp{R}}_{Q,i}]_{k k}^\top [ \M \BSigma \M ]_{k k}   [\cb{\cp{R}}_{Q(I-H),i}]_{k \ell} \}      \end{align}
 \normalsize
 Replacing $[\cb{\cp{R}}_{Q,i}]_{k k}$ and $[\cb{\cp{R}}_{Q(I-H),i}]$ by their definitions~\eqref{eq:R_Q} and~\eqref{eq:R_Q_I_H}, respectively, we get:
 \begin{align}
     &  [\bP_3]_{k \ell} = \\
     & \sum_{m=1}^N c_{mk} c_{\ell k}  \E \{\bH_{k,i}  \bR_{u_m,i} \bQ_{m,i} [ \M \BSigma \M ]_{k k} \nonumber \\
     & \hspace{4cm} \bQ_{\ell ,i}  \bR_{u_\ell,i} \left(  \I_L - \bH_{k,i} \right) \}   \nonumber
     \end{align}
 Applying \eqref{eq:mean_H} and \eqref{eq:cov_Q_ll} leads to:
 \begin{align}    
     &  [\bP_3]_{k \ell} =  \frac{M}{L} \, c_{\ell k}^2 \, \E \{\bR_{u_\ell}  \bQ_{\ell,i} [ \M \BSigma \M ]_{k k}   \bQ_{\ell,i} \bR_{u_\ell}  \} \nonumber \\
     & \hspace{5mm} -  c_{\ell k}^2 \, \E \{ \bH_{k,i}   \bR_{u_\ell}  \bQ_{\ell,i} [ \M \BSigma \M ]_{k k}    \bQ_{\ell,i}  \bR_{u_\ell} \bH_{k,i} \}  \nonumber \\ 
     & \hspace{5mm} +  \left( \frac{\Mg}{L} \right)^2 \Big( \frac{M}{L}  \sum_{m=1}^N c_{mk} c_{\ell k}  \bR_{u_m}  [ \M \BSigma \M ]_{k k}   \bR_{u_\ell}  \nonumber \\
     & \hspace{5mm} -  \sum_{m=1}^N c_{mk} c_{\ell k} \E \{ \bH_{k,i} \bR_{u_m}  [ \M \BSigma \M ]_{k k}   \bR_{u_\ell} \bH_{k,i} \} \nonumber \\
     & \hspace{5mm} - \frac{M}{L} \,  c_{\ell k}^2  \,  \bR_{u_\ell}  [ \M \BSigma \M ]_{k k}  \bR_{u_\ell}  \nonumber \\
     & \hspace{5mm} +  c_{\ell k}^2 \, \E \{ \bH_{k,i} \bR_{u_\ell}  [ \M \BSigma \M ]_{k k}  \bR_{u_\ell} \bH_{k,i} \}   \Big)
    \label{eq:P_3}
     \end{align}
     \normalsize
     We have:
     \begin{equation}
         c_{\ell k}^2 \, \bR_{u_\ell} [ \M \BSigma \M ]_{k k}\bR_{u_\ell}
           = \sum_{m=1}^N [\cpb{R}'_{u_m} \M \BSigma \M \C_2^\top\cpb{I }_m
         \cpb{R}_u ]_{k \ell}
         \label{eq:reform_1}
     \end{equation}
     where
     \begin{align}
              & \cb{\cp{R}}'_{u_m} = \I_L \otimes \bR_{u_m} \label{eq:R_Tm_def} \\
          & \cb{\cp{I}}_m = \diag \{  0 , 0 , \dots , \I_L , 0, \dots , 0 \} 
   \end{align}
     
     All the entries of the matrix $ \cb{\cp{I}}_m $ are equal to zero except the $(m,m)$-th block which is equal to $ \I_L $.
     
     Using \eqref{eq:reform_1}, we find that:
          \begin{align}
     \bP_3  & =  \frac{M}{L} \sum_{m=1}^N \,  \cpb{R}'_{u_m}  \bvphi_Q( \M \BSigma \M ) \C_2^\top \cpb{I }_m  \cpb{R}_{u}  \nonumber \\
    &-   \sum_{m=1}^N \,  \bvphi_H( \cpb{R}'_{u_m}   \bvphi_Q(\M \BSigma \M)  \C_2^\top  \cpb{I }_m \cpb{R}_{u} )  \nonumber \\ 
    &+  \left( \frac{\Mg}{L} \right)^2 \Bigg( \frac{M}{L}  
     \cpb{R}  \M \BSigma \M  \C^\top   \cpb{R}_{u}  
    -  \bvphi_H( \cpb{R}  \M \BSigma \M  \C^\top   \cpb{R}_{u} ) \nonumber \\
    &-  \frac{M}{L} \, \sum_{m=1}^N \,  \cpb{R}'_{u_m} \M \BSigma \M  \C_2^\top \cpb{I }_m  \cpb{R}_{u}  \nonumber \\
    &+  \sum_{m=1}^N \, \bvphi_H( \cpb{R}'_{u_m}   \M \BSigma \M \C_2^\top  \cpb{I }_m \cpb{R}_{u} ) \Bigg)
     \end{align}
                                    \subsubsection{ Term $\bP_4$ calculation }
 
 We express $\bP_4$ as follows:
     \begin{align}
     [\bP_4]_{k k} & =  \E \{  \bR_{u_k,i} [ \cpb{Q}_{i}']_{kk} [ \M \BSigma \M ]_{k k} [ \cpb{Q}_{i}']_{kk} \bR_{u_k,i}   \}   \nonumber 
     \end{align}
     Substituting $ [ \cpb{Q}_{i}']_{kk} $ by its definition \eqref{eq:Qp_def} we get:
     \begin{align}
     &  [\bP_4]_{k k} = \sum_{m,n=1}^N c_{mk} c_{nk}  \E \{  \bR_{u_k,i} \left( \I_L - \bQ_{m,i} \right)  [ \M \BSigma \M ]_{k k} \nonumber \\
     & \hspace{4cm}  \left( \I_L - \bQ_{n,i} \right) \bR_{u_k,i}   \}   \nonumber 
     \end{align}
     Using \eqref{eq:mean_H} leads to 
          \begin{align}
     [\bP_4]_{k k} &=  \left (  1 - 2  \frac{\Mg}{L} \right) \sum_{m,n=1}^N c_{mk} c_{nk}  \E \{  \bR_{u_k,i}   [ \M \BSigma \M ]_{k k}  \bR_{u_k,i}   \}  \nonumber \\
     + &  \sum_{m,n=1}^N c_{mk} c_{nk}  \E \{   \bR_{u_k,i}  \bQ_{m,i}  [ \M \BSigma \M ]_{k k}  \bQ_{n,i}  \bR_{u_k,i}   \}  \nonumber 
     \end{align}
           We rearrange the sum as follows:
          \begin{align}
      [\bP_4]_{k k} & =  \left (  1 - 2  \frac{\Mg}{L} \right) \sum_{m,n=1}^N c_{mk} c_{nk}  \E \{   \bR_{u_k,i}   [ \M \BSigma \M ]_{k k}  \bR_{u_k,i}   \}  \nonumber \\
     + &  \sum_{m=1}^N c_{mk}^2  \E \{   \bR_{u_k,i}  \bQ_{m,i}  [ \M \BSigma \M ]_{k k}  \bQ_{m,i}  \bR_{u_k,i}   \}  \nonumber \\
     + & \left( \frac{\Mg}{L} \right)^2  \left ( \sum_{m,n=1}^N c_{mk} c_{nk}  \E \{   \bR_{u_k,i}   [ \M \BSigma \M ]_{k k}    \bR_{u_k,i}  \} \right. \nonumber \\
     -  & \left. \sum_{m=1}^N c_{mk}^2  \E \{  \bR_{u_k,i}   [ \M \BSigma \M ]_{k k}    \bR_{u_k,i}  \} \right)  \nonumber 
      \end{align}
          Finally, we can write $ \bP_4 $ in a compact form:
     \begin{align}
     &\bP_4   =  \left (  1 - 2  \frac{\Mg}{L} \right)   \cpb{R}_{u}     \M \BSigma \M   \cpb{R}_{u} \nonumber  \\
     &  \hspace{1cm}+   \cpb{R}_{u}  (\I_{NM} \odot \C \, \C^\top ) \, \bvphi_Q(  \M \BSigma \M   )  \cpb{R}_{u}  \nonumber \\
     &  \hspace{1cm}+  \left( \frac{\Mg}{L} \right)^2  \left (   \cpb{R}_{u}    \M \BSigma \M    \cpb{R}_{u}  \right. \nonumber \\
     &  \hspace{1cm} \left. -    \cpb{R}_{u} (\I_{NM} \odot \C \, \C^\top )    \M \BSigma \M     \cpb{R}_{u} \right)
     \end{align}
 
 \subsubsection{ Term $\bP_5 $ calculation}
 
     Expanding $ \bP_5 $ leads to:
     \begin{equation}
         \bP_5 = \bP_{5,1} - \bP_{5,2} - \bP_{5,3} + \bP_{5,4} 
     \label{eq:P_6t}
     \end{equation}
     where:
                  \begin{align}
         & \bP_{5,1}  =  \sum_{m=1}^N c_{mk} c_{ \ell k}  \E \{  \bR_{u_k,i}  [ \M \BSigma \M ]_{k k} \bQ_{\ell,i} \bR_{u_\ell,i} \}  \\
         & \bP_{5,2}   =\sum_{m=1}^N c_{mk} c_{ \ell k}  \E \{  \bR_{u_k,i} \bQ_{m,i}  [ \M \BSigma \M ]_{k k} \nonumber \\  
         & \hspace{6cm} \bQ_{\ell,i} \bR_{u_\ell,i}  \} \\
         & \bP_{5,3}  = \sum_{m=1}^N c_{mk} c_{ \ell k}  \E \{  \bR_{u_k,i}   [ \M \BSigma \M ]_{k k} \nonumber \\
         & \hspace{5cm} \bQ_{\ell,i} \bR_{u_\ell,i} \bH_{k,i}  \} \\
         &\bP_{5,4}  = \sum_{m=1}^N c_{mk} c_{ \ell k}  \E \{  \bR_{u_k,i}  \bQ_{m,i}  [ \M \BSigma \M ]_{k k} \nonumber \\
         & \hspace{5cm} \bQ_{\ell,i} \bR_{u_\ell,i} \bH_{k,i} \} 
     \end{align}
      Following the same steps as earlier, and using the results \eqref{eq:mean_H} and \eqref{eq:cov_Q_ll}, leads to:
             \begin{align}
         & \bP_{5,1}  = \frac{\Mg}{L} \cpb{R}_{u} \M \BSigma \M \C^\top \cpb{R}_{u}  \\
         & \bP_{5,2}   = \frac{M}{L} \Big[ \cpb{R}_{u} \bvphi_Q (\M \BSigma \M) \C_2^\top \cpb{R}_{u}  \\
         &+  \left( \frac{\Mg}{L} \right)^2 \left( \cpb{R}_{u} \M \BSigma \M \C^\top \cpb{R}_{u}    
          - \cpb{R}_{u}  \M \BSigma \M \C_2^\top \cpb{R}_{u}  \right)  \Big]   \nonumber \\
         & \bP_{5,3}  =\frac{M}{L}  \cpb{R}_{u} \M \BSigma \M \C^\top \cpb{R}_{u} \\
         &\bP_{5,4}  = \frac{M}{L}  \bvphi_Q ( \M \BSigma \M) \C_2^\top \bR_{u}   \\
         &+\frac{M}{L} \left( \frac{\Mg}{L} \right)^2 \left(  \cpb{R}_{u} \M \BSigma \M \C^\top \bR_{u}
         -  \cpb{R}_{u}   \M \BSigma \M \C_2^\top \bR_{u} \right) 	\nonumber	
     \end{align}
      \subsubsection{ Term $\bP_6 $ calculation}
 
     Proceeding as previously we find:
          \begin{align}
     & \bP_6 = \left( 1 -  \frac{2M}{L} \right)  \cpb{R}_{u} \E \{ \cpb{Q}_i \C \M \BSigma \M \C^\top \cpb{Q}_i\}  \cpb{R}_{u}  \nonumber \\
     & \hspace{1.2cm} +   \bvphi_H( \cpb{R}_{u} \E \{ \cpb{Q}_i \C \M \BSigma \M \C^\top \cpb{Q}_i\}   \cpb{R}_{u} )
     \end{align}
      
 \subsection{Terms $\bP_j$ vectorization}
 In order to apply the $\vc(\cdot)$ operator to the terms $\bP_j$, we use the following transformations:
 
        \begin{align}
         &(\I_{N} \otimes \boldsymbol{1}_{LL}) \odot \cb{\Pi} = \sum_{n=1}^N \bT_n \cb{\Pi} \bT_n \label{eq:blk_diag}  \\
          &\I_{N L} \odot \cb{\Pi} =  \sum_{n=1}^{NL} \bD_n \cb{\Pi} \bD_n  \label{eq:diag} \\
         &( \boldsymbol{1}_{NN} \otimes  \I_{L}) \odot \cb{\Pi} = \nonumber \\ 
         &\hspace{2cm}\sum_{m=1}^{NL} \sum_{k=1}^{\left \lfloor{\frac{NL-M}{L}}\right \rfloor} (\bD_m \cb{\Pi} \bD_{m+kL}  + \bD_{m+kL} \cb{\Pi} \bD_m) \nonumber \\ 
     & \hspace{2cm} - \sum_{m=1}^N \bD_m \cb{\Pi} \bD_m
     \end{align}
     where $\cb{\Pi}$, $\bT_n$ and $\bD_n$, are $(NL \times NL)$ matrices and 
     \begin{align}
         &[\bT_n]_{k \ell} = \delta(k, \ell) \delta(k,n) \I_L \\
         &(\bD_n)_{k \ell}= \delta(k, \ell) \delta(k,n) 
     \end{align}
     where $\delta(k, \ell)=1$ if $k=\ell$, and $0$ otherwise. 
     Using \eqref{eq:EQ}, we find:
     \begin{align}
    & \vc(\cpb{R}_{u} \E \{ \cpb{Q}_i \C \M \BSigma \M \C^\top \cpb{Q}_i \}  \cpb{R}_{u})= \nonumber \\
    & \hspace{1cm}\alpha_1 \, \vc(  \cpb{R}_{u}[( \I_{N} \otimes \boldsymbol{1}_{LL}) \odot (\C \M \BSigma \M \C^\top)]\cpb{R}_{u}) \nonumber \\
    & \hspace{5mm}+ \alpha_2 \, \vc(  \cpb{R}_{u}[ \I_{N}   \odot (\C \M \BSigma \M \C^\top)]\cpb{R}_{u}) \nonumber \\
    & \hspace{5mm}+ \alpha_3 \, \vc(  \cpb{R}_{u}  \C \M \BSigma \M \C^\top  \cpb{R}_{u})
     \end{align}
     Using \eqref{eq:blk_diag} and \eqref{eq:diag} we have:
         \begin{align}
    & \vc(\cpb{R}_{u} \E \{ \cpb{Q}_i \C \M \BSigma \M \C^\top \cpb{Q}_i \}  \cpb{R}_{u})= \nonumber \\
    & \hspace{1cm}\alpha_1 \, \sum_{n=1}^{N} \vc(  \cpb{R}_{u} \bT_n  \C \M \BSigma \M \C^\top \bT_n  \cpb{R}_{u}) \nonumber \\   
    & \hspace{5mm}+ \alpha_2 \, \sum_{m=1}^{NL} \vc(  \cpb{R}_{u} \bD_m \C \M \BSigma \M \C^\top \bD_m \cpb{R}_{u}) \nonumber \\
    & \hspace{5mm}+ \alpha_3 \, \vc(  \cpb{R}_{u}  \C \M \BSigma \M \C^\top  \cpb{R}_{u})
     \end{align}
     Furthermore, we have:
 \begin{equation}  
     \vc (\bA \bSigma \bB) = (\bB^\top\! \otimes \bA) \vc(\bSigma) \label{eq:vec_prop}
 \end{equation}
 Applying \eqref{eq:vec_prop} leads to:
         \begin{align}
    & \vc(\cpb{R}_{u} \E \{\cpb{Q}_i  \C \M \BSigma \M \C^\top \cpb{Q}_i \}  \cpb{R}_{u})= \nonumber \\
    & \hspace{1cm}\alpha_1 \, \sum_{n=1}^{N} \vc( \cpb{R}_{u} \bT_n \C \M \otimes  \cpb{R}_{u} \bT_n  \C \M) \bsigma    \nonumber \\
    & \hspace{5mm}+ \alpha_2 \, \sum_{m=1}^{NL} \vc( \cpb{R}_{u} \bD_m \C \M \otimes  \cpb{R}_{u} \bD_m  \C \M) \bsigma    \nonumber \\
    & \hspace{5mm}+ \alpha_3 \, \vc( \cpb{R}_{u}  \C \M \otimes  \cpb{R}_{u}   \C \M) \bsigma    
     \end{align}
   
 \bibliographystyle{IEEEtran}

\begin{thebibliography}{10}
\providecommand{\url}[1]{#1}
\csname url@samestyle\endcsname
\providecommand{\newblock}{\relax}
\providecommand{\bibinfo}[2]{#2}
\providecommand{\BIBentrySTDinterwordspacing}{\spaceskip=0pt\relax}
\providecommand{\BIBentryALTinterwordstretchfactor}{4}
\providecommand{\BIBentryALTinterwordspacing}{\spaceskip=\fontdimen2\font plus
\BIBentryALTinterwordstretchfactor\fontdimen3\font minus
  \fontdimen4\font\relax}
\providecommand{\BIBforeignlanguage}[2]{{%
\expandafter\ifx\csname l@#1\endcsname\relax
\typeout{** WARNING: IEEEtran.bst: No hyphenation pattern has been}%
\typeout{** loaded for the language `#1'. Using the pattern for}%
\typeout{** the default language instead.}%
\else
\language=\csname l@#1\endcsname
\fi
#2}}
\providecommand{\BIBdecl}{\relax}
\BIBdecl

\bibitem{sayed2013difadapt}
A.~H. Sayed, ``Diffusion adaptation over networks,'' in \emph{Academic Press
  Libraray in Signal Processing}, R.~Chellapa and S.~Theodoridis, Eds.\hskip
  1em plus 0.5em minus 0.4em\relax Elsevier, 2014. Also available as
  arXiv:1205.4220 [cs.MA], May 2012, pp. 322--454.

\bibitem{nedic2009distributed}
A.~Nedic and A.~Ozdaglar, ``Distributed subgradient methods for multi-agent
  optimization,'' \emph{IEEE Transactions on Automatic Control}, vol.~54,
  no.~1, pp. 48--61, 2009.

\bibitem{rabbat2005quantized}
M.~G. Rabbat and R.~D. Nowak, ``Quantized incremental algorithms for
  distributed optimization,'' \emph{IEEE Journal on Selected Areas in
  Communications}, vol.~23, no.~4, pp. 798--808, 2005.

\bibitem{lopes2007incremental}
C.~G. Lopes and A.~H. Sayed, ``Incremental adaptive strategies over distributed
  networks,'' \emph{IEEE Transactions on Signal Processing}, vol.~55, no.~8,
  pp. 4064--4077, 2007.

\bibitem{sayed2013diffusion}
A.~H. Sayed, S.-Y. Tu, J.~Chen, X.~Zhao, and Z.~J. Towfic, ``Diffusion
  strategies for adaptation and learning over networks: an examination of
  distributed strategies and network behavior,'' \emph{IEEE Signal Processing
  Magazine}, vol.~30, no.~3, pp. 155--171, 2013.

\bibitem{sayed2014adaptive}
A.~H. Sayed, ``Adaptive networks,'' \emph{Proceedings of the IEEE}, vol. 102,
  no.~4, pp. 460--497, 2014.

\bibitem{chen2014diffusion3}
J.~Chen, C.~Richard, A.~O. Hero, and A.~H. Sayed, ``Diffusion {LMS} for
  multitask problems with overlapping hypothesis subspaces,'' in \emph{Proc.
  IEEE MLSP'14}, Reims, France, 2014, pp. 1--6.

\bibitem{chen2014multitask}
J.~Chen, C.~Richard, and A.~H. Sayed, ``Multitask diffusion adaptation over
  networks,'' \emph{IEEE Transactions on Signal Processing}, vol.~62, no.~16,
  pp. 4129--4144, 2014.

\bibitem{chen2015diffusion}
------, ``Diffusion {LMS} over multitask networks,'' \emph{IEEE Transactions on
  Signal Processing}, vol.~63, no.~11, pp. 2733--2748, 2015.

\bibitem{nassif2015multitask1}
R.~Nassif, C.~Richard, A.~Ferrari, and A.~H. Sayed, ``Multitask diffusion
  adaptation over asynchronous networks,'' \emph{IEEE Transactions on Signal
  Processing}, vol.~64, no.~11, pp. 2835--2850, 2016.

\bibitem{nassif2015proximal}
------, ``Proximal multitask learning over networks with sparsity-inducing
  coregularization,'' \emph{IEEE Transactions on Signal Processing}, vol.~64,
  no.~23, pp. 6329--6344, 2016.

\bibitem{chen2016multitask}
J.~Chen, C.~Richard, and A.~H. Sayed, ``Multitask diffusion adaptation over
  networks with common latent representations,'' \emph{IEEE Journal of Selected
  Topics in Signal Processing}, vol.~11, no.~3, pp. 563--579, 2017.

\bibitem{takahashi2010diffusion}
N.~Takahashi, I.~Yamada, and A.~H. Sayed, ``Diffusion least-mean squares with
  adaptive combiners: Formulation and performance analysis,'' \emph{IEEE
  Transactions on Signal Processing}, vol.~58, no.~9, pp. 4795--4810, 2010.

\bibitem{Khalili2012}
A.~Khalili, M.~A. Tinati, A.~Rastegarnia, and J.~A. Chambers, ``Steady-state
  analysis of diffusion {LMS} adaptive networks with noisy links,''
  \emph{{IEEE} Transaction on Signal Processing}, vol.~60, no.~2, pp. 974--979,
  2012.

\bibitem{Zhao2012impe}
X.~Zhao, S.-Y. Tu, and A.~H. Sayed, ``Diffusion adaptation over networks under
  imperfect information exchange and non-stationary data,'' \emph{{IEEE}
  Transaction on Signal Processing}, vol.~60, no.~7, pp. 3460--3475, 2012.

\bibitem{ChenUCLA2012}
J.~Chen and A.~H. Sayed, ``Diffusion adaptation strategies for distributed
  optimization and learning over networks,'' \emph{{IEEE} Transactions on
  Signal Processing}, vol.~60, no.~8, pp. 4289--4305, 2012.

\bibitem{ChenUCLA2013}
------, ``Distributed {P}areto optimization via diffusion strategies,''
  \emph{IEEE Journal of Selected Topics in Signal Processing}, vol.~7, no.~2,
  pp. 205--220, 2013.

\bibitem{Gharehshiran2013}
O.~N. Gharehshiran, V.~Krishnamurthy, and G.~Yin, ``Distributed energy-aware
  diffusion least mean squares: Game-theoretic learning,'' \emph{IEEE Journal
  of Selected Topics in Signal Processing}, vol.~7, no.~5, pp. 1--16, 2013.

\bibitem{Chouvardas2011set}
S.~Chouvardas, K.~Slavakis, and S.~Theodoridis, ``Adaptive robust distributed
  learning in diffusion sensor networks,'' \emph{{IEEE} Transactions on Signal
  Processing}, vol.~59, no.~10, pp. 4692--4707, 2011.

\bibitem{sayed2014adaptation}
A.~H. Sayed, ``Adaptation, learning, and optimization over networks,'' in
  \emph{Foundations and Trends in Machine Learning}.\hskip 1em plus 0.5em minus
  0.4em\relax Boston-Delft: NOW Publishers, 2014, vol.~7, no. 4-5, pp.
  311--801.

\bibitem{Liu2012}
Y.~Liu, C.~Li, and Z.~Zhang, ``Diffusion sparse least-mean squares over
  networks,'' \emph{{IEEE} Transactions on Signal Processing}, vol.~60, no.~8,
  pp. 4480--4485, 2012.

\bibitem{Chouvardas2012}
S.~Chouvardas, K.~Slavakis, Y.~Kopsinis, and S.~Theodoridis, ``A
  sparsity-promoting adaptive algorithm for distributed learning,''
  \emph{{IEEE} Transactions on Signal Processing}, vol.~60, no.~10, pp.
  5412--5425, 2012.

\bibitem{Lorenzo2013spar}
P.~Di~Lorenzo and A.~H. Sayed, ``Sparse distributed learning based on diffusion
  adaptation,'' \emph{{IEEE} Transactions on Signal Processing}, vol.~61,
  no.~6, pp. 1419--1433, 2013.

\bibitem{wen2015diffusion}
F.~Wen and W.~Liu, ``Diffusion least mean square algorithms with
  zero-attracting adaptive combiners,'' in \emph{Proc. IEEE
  CIT--IUCC--DASC--PICOM'15}, 2015, pp. 252--256.

\bibitem{Predd2006WSN}
J.~Predd, S.~Kulkarni, and H.~Vincent~Poor, ``Distributed learning in wireless
  sensor networks,'' \emph{IEEE Signal Processing Magazine}, vol.~23, no.~4,
  pp. 59--69, 2006.

\bibitem{chainais2013learning}
P.~Chainais and C.~Richard, ``Learning a common dictionary over a sensor
  network,'' in \emph{Proc. IEEE Int. Workshop on Computational Advances in
  Multi-Sensor Adaptive Processing (CAMSAP)}, Saint Martin, France, Dec. 2013,
  pp. 1--4.

\bibitem{gao2015diffusion}
W.~Gao, J.~Chen, C.~Richard, and J.~Huang, ``Diffusion adaptation over networks
  with kernel least-mean-square,'' in \emph{Proc. IEEE CAMSAP'15}, Canc\'un,
  Mexico, 2015, pp. 217--220.

\bibitem{lopes2008diffusion}
C.~G. Lopes and A.~H. Sayed, ``Diffusion adaptive networks with changing
  topologies,'' in \emph{Proc. IEEE ICASSP'08}, Las Vegas, USA, 2008, pp.
  3285--3288.

\bibitem{arablouei2015analysis}
R.~Arablouei, S.~Werner, K.~Do{\u{g}}an{\c{c}}ay, and Y.-F. Huang, ``Analysis
  of a reduced-communication diffusion {LMS} algorithm,'' \emph{Signal
  Processing}, vol. 117, pp. 355--361, 2015.

\bibitem{sayin2014compressive}
M.~O. Sayin and S.~S. Kozat, ``Compressive diffusion strategies over
  distributed networks for reduced communication load,'' \emph{IEEE
  Transactions on Signal Processing}, vol.~62, no.~20, pp. 5308--5323, 2014.

\bibitem{arablouei2014distributed}
R.~Arablouei, S.~Werner, Y.-F. Huang, and K.~Do{\u{g}}an{\c{c}}ay,
  ``Distributed least mean-square estimation with partial diffusion,''
  \emph{IEEE Transactions on Signal Processing}, vol.~62, no.~2, pp. 472--484,
  2014.

\bibitem{arablouei2014adaptive}
R.~Arablouei, K.~Do{\u{g}}an{\c{c}}ay, S.~Werner, and Y.-F. Huang, ``Adaptive
  distributed estimation based on recursive least-squares and partial
  diffusion,'' \emph{IEEE Transactions on Signal Processing}, vol.~62, no.~14,
  pp. 3510--3522, 2014.

\bibitem{vadidpour2015partial}
V.~Vadidpour, A.~Rastegarnia, A.~Khalili, and S.~Sanei, ``Partial-diffusion
  least mean-square estimation over networks under noisy information
  exchange,'' \emph{arXiv preprint arXiv:1511.09044}, 2015.

\bibitem{necoara2017random}
I.~Necoara, Y.~Nesterov, and F.~Glineur, ``Random block coordinate descent
  methods for linearly constrained optimization over networks,'' \emph{Journal
  of Optimization Theory and Applications}, vol. 173, no.~1, pp. 227--254,
  2017.

\bibitem{xi2017distributed}
C.~Xi and U.~A. Khan, ``Distributed subgradient projection algorithm over
  directed graphs,'' \emph{IEEE Transactions on Automatic Control}, vol.~62,
  no.~8, pp. 3986--3992, 2017.

\bibitem{wang2018coordinate}
C.~Wang, Y.~Zhang, B.~Ying, and A.~H. Sayed, ``Coordinate-descent diffusion
  learning by networked agents,'' \emph{IEEE Transactions on Signal
  Processing}, vol.~66, no.~2, pp. 352--367, 2018.

\bibitem{le2013multi}
T.~N. Le, A.~Pegatoquet, O.~Berder, and O.~Sentieys, ``Multi-source power
  manager for super-capacitor based energy harvesting {WSN},'' in \emph{Proc.
  ACM ENSSys'13}, Rome, Italy, 2013, pp. 19:1--19:2.

\end{thebibliography}

\balance

\end{document}